\documentclass{article}

\usepackage{microtype}
\usepackage{graphicx}
\usepackage{subfigure}
\usepackage{booktabs} %

\usepackage{hyperref}

\usepackage[accepted]{icml2023}

\usepackage{amsmath}
\usepackage{amssymb}
\usepackage{mathtools}
\usepackage{amsthm}
\usepackage{bbm}
\usepackage[capitalize,noabbrev]{cleveref}
\usepackage{cancel}
\usepackage{makecell}

\theoremstyle{plain}
\newtheorem{theorem}{Theorem}[section]

\theoremstyle{definition}

\theoremstyle{remark}

\usepackage[textsize=tiny]{todonotes}

\usepackage{nkj}

\icmltitlerunning{Relevant Walk Search for Explaining Graph Neural Networks}

\begin{document}

\twocolumn[
\icmltitle{Relevant Walk Search for Explaining Graph Neural Networks}

\icmlsetsymbol{equal}{*}

\begin{icmlauthorlist}

\icmlauthor{Ping Xiong}{tu,bf}
\icmlauthor{Thomas Schnake }{tu,bf}
\icmlauthor{Michael Gastegger }{tu}
\icmlauthor{Gr\'{e}goire Montavon}{fu,bf}
\icmlauthor{Klaus-Robert M\"{u}ller}{tu,bf,korea,saar,google}
\icmlauthor{Shinichi Nakajima}{tu,bf,riken}

\end{icmlauthorlist}

\icmlaffiliation{tu}{Technische Universit\"{a}t Berlin (TU Berlin)}
\icmlaffiliation{fu}{Freie Universit\"{a}t Berlin (FU Berlin)}
\icmlaffiliation{bf}{BIFOLD -- Berlin  Institute  for  the Foundations  of  Learning  and  Data}
\icmlaffiliation{google}{Google Research, Brain team, Berlin}
\icmlaffiliation{korea}{Department of Artificial Intelligence, Korea University, Seoul 136-713, Korea}
\icmlaffiliation{saar}{Max Planck Institut für Informatik, 66123 Saarbrücken, Germany}
\icmlaffiliation{riken}{RIKEN Center for AIP, Japan}

\icmlcorrespondingauthor{Shinichi Nakajima}{nakajima@tu-berlin.de}

\icmlkeywords{graph neural networks, XAI, layer-wise relevance propagation, message passing}

\vskip 0.3in
]

\printAffiliationsAndNotice{}  %

\begin{abstract}
Graph Neural Networks (GNNs) have become important machine learning tools for graph analysis, and its explainability is crucial for safety, fairness, and robustness.  Layer-wise relevance propagation for GNNs (GNN-LRP) evaluates the relevance of \emph{walks} to reveal important information flows in the network, and provides higher-order explanations, which have been shown to be superior to the lower-order, i.e., node-/edge-level, explanations.  However, identifying relevant walks by GNN-LRP requires {\em exponential} computational complexity with respect to the network depth, which we will remedy in this paper.  Specifically, we propose {\em polynomial-time} algorithms for finding top-$K$ relevant walks, which drastically reduces the computation and thus increases the applicability of GNN-LRP to large-scale problems.  Our proposed algorithms are based on the \emph{max-product} algorithm---a common tool for finding the maximum likelihood configurations in probabilistic graphical models---and can find the most relevant walks exactly at the neuron level and approximately at the node level.  Our experiments demonstrate the performance of our algorithms at scale and their utility across application domains, i.e., on epidemiology, molecular, and natural language benchmarks. We provide our codes under \href{https://github.com/xiong-ping/rel_walk_gnnlrp}{github.com/xiong-ping/rel\_walk\_gnnlrp}.
\end{abstract}

\section{Introduction}
\label{sec:Introduction}

Graph Neural Networks (GNNs) are powerful machine learning tools to solve tasks on graph datasets, such as social networks \citep{DBLP:conf/kdd/YanardagV15} and molecules \citep{doi:10.1021/jm040835a, schutt2018schnet}. 
Various GNN models have been proposed \citep{DBLP:journals/tnn/WuPCLZY21}, while the rationale of 
prediction---for example, which node features  or which parts of the input graph are (jointly) contributing to the prediction---cannot be directly extracted from the model. The GNN therefore acts as a black box and its prediction is hard to comprehend without further investigation. 
To address this challenge,
many explainability methods emerged in recent years \citep{schnake2022higher, ying2019gnnexplainer},
which can be generally categorized into model-level and instance-level methods \citep{yuan2020explainability}: model-level methods generate representative graphs for different predictions, while instance-level methods focus on single predictions and find  relevant features in the corresponding input graphs.

\begin{figure*}
    \centering
    \vspace{-3mm}    \includegraphics[width=.86\textwidth]{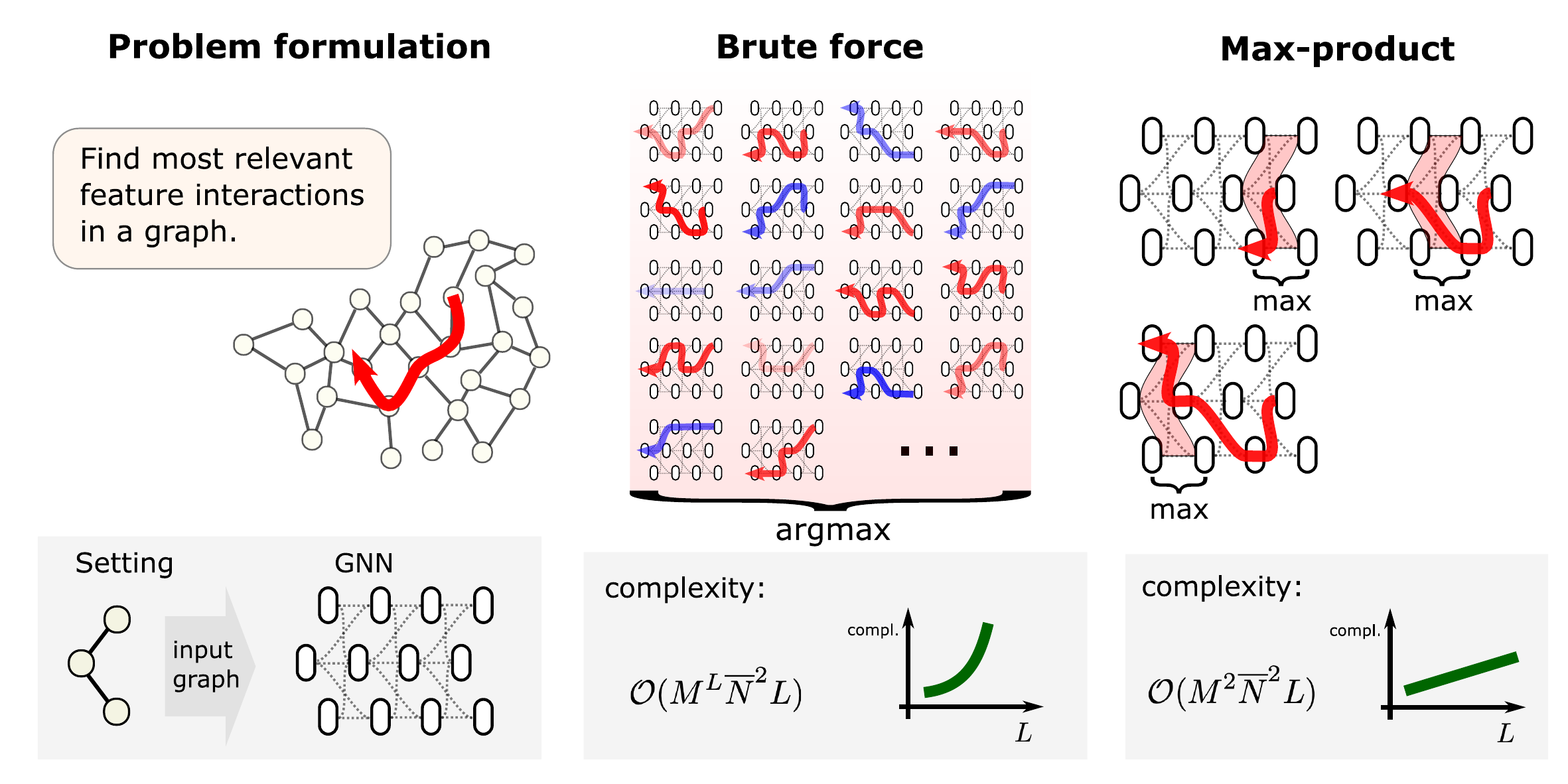}

    \vspace{-4mm}
    
    \caption{
        We aim to find the most important information flows for a GNN prediction in terms of walk relevance.
       Naively one can apply a brute force search, where the relevances of all possible walks are evaluated and the most relevant ones are chosen.
       Our proposed methods are based on the max-product algorithm to find the most relevant walks by local message passing, which reduces the computational complexity drastically from exponential to linear with respect to the network depth.
    }
    \label{fig:fig2}
    \vspace{-3mm}
\end{figure*}

This paper focuses on the instance-level explanation,
for which most of the existing methods, e.g., GNNExplainer \citep{ying2019gnnexplainer}, PGExplainer \cite{luo2020parameterized} and methods in \citet{DBLP:conf/cvpr/PopeKRMH19},  only consider lower-order features, i.e., nodes and edges, ignoring higher-order interactions.
Recently, \citet{schnake2022higher}
proposed layer-wise relevance propagation for GNNs (GNN-LRP) as a decomposition of the model into contributions of input features that jointly give rise to the prediction. In particular, this leads us to a higher-order explanation method that measures the relevance of \emph{walks} on the input graph.
GNN-LRP can be used in two ways: (1) identifying relevant walks that reveal important information flows in the network,
and (2) aggregating walk relevances within subgraphs or substructures in the input graph to attribute them.
In either way, GNN-LRP 
captures interactions between features across the network layers, and has shown its superior performance to the lower-order methods.

Despite its ability to enable 
higher-order explanation, the GNN-LRP %
algorithm faces computational issues: both its relevant walk identification and the subgraph attribution require exponentially many walks to be evaluated. 
Therefore naive implementations can cope only with small-scale problems in terms of the network depth and the graph size.  
The latter issue, namely exponential complexity for subgraph attribution, was 
solved by \citet{DBLP:conf/icml/XiongSMMN22}.
They pointed out that the relevance of a walk has the same decomposability to the joint distribution of a Markov chain process, and applied the \emph{sum-product} message passing algorithm for computing the subgraph relevance, which reduces the computational cost from exponential to polynomial.
They also showed that the general LRP computation for feed-forward neural networks can be seen as marginalization of walk relevances by the sum-product message passing algorithm.

In this paper, we tackle the other issue: exponential complexity for relevant walk identification.  
First of all, we would like to propose 
 that the decomposability argument in \citet{DBLP:conf/icml/XiongSMMN22} could also invoke the applicability of the \emph{max-product} message passing algorithm. This idea could allow us the usage of a popular workhorse for finding the maximum likelihood configuration of probabilistic graphical models \citep{Viterbi67,bishop2006patternchp8}.
Our proposal indeed relies on the max-product decomposability of the objective, and thus allows deriving local message passing for relevance maximization
(see Fig.\ref{fig:fig2}).
However, two difficulties stand in our way by obstructing max-product decompositions: (1) negative values that relevance factors can take, and (2) sum operations involved in the node-level walk search.%
\footnote{
The neuron- and node-level walks (defined in \citet{DBLP:conf/icml/XiongSMMN22})
will be explained in \cref{sec:WalkRelevance},
and 
illustrated in Fig.\ref{fig:NeuronNodeLevelWalks}.
}
We resolve those difficulties by finding multiple \emph{absolute} relevant walks with search space splitting \citep{nilsson1998efficient} and approximation with neuron averaging,
and propose two algorithms in this paper: an \underline{\textbf{e}}xact \underline{\textbf{m}}ax-\underline{\textbf{p}}roduct search for \underline{\textbf{neu}}ron-level walks (EMP-neu), and an \underline{\textbf{a}}pproximate \underline{\textbf{m}}ax-\underline{\textbf{p}}roduct search by \underline{\textbf{ave}}raging (AMP-ave) for node-level walks.

The computational complexities of EMP-neu and AMP-ave are polynomial, so that we obtain an explanation method that has the same higher-order feature resolution as GNN-LRP but without exponential computational costs. In addition we provide a variety of  qualitative and quantitative experiments
that demonstrate the usefulness of our approach. We show as well that the accuracy of AMP-ave is high in our experiments, so that the potential approximation error is negligible throughout the domains. This leads us to a new explanation method, with only an insignificant approximation error, that is fast computable and exhibits   superior higher-order feature resolution.

\section{Background and Related Works}

\subsection{Graph Neural Networks}
\label{sec:GraphNeuralNetworks}

Graph Neural Networks (GNNs) \citep{DBLP:journals/tnn/ScarselliGTHM09, wu2020comprehensive} 
take a graph as an input, and make the corresponding prediction by using the topological structure of the graph.
In our setting we consider the message passing neural networks (MPNNs) \citep{gilmer2017neural}, which learn node embeddings in multiple interaction blocks, and each block 
typically consists of \emph{aggregate} and \emph{combine} steps:
\begin{align}
        \text{Aggregate: } \boldsymbol Z^{(l)} &= \mathcal{M}^{(l)}(  \boldsymbol{H}^{(l-1)}, \bfLambda), 
        \label{eq:GNN_agg}\\
        \text{Combine: } \boldsymbol H^{(l)} &= \textstyle \mathcal C^{(l)}\left(\boldsymbol Z^{(l)}
        \right).
     \label{eq:GNN_comb}
\end{align}
Here $\boldsymbol H^{(l)} \in \mathbb R^{M\times N^{(l)}}$ is the feature (activation) matrix of the $l$-th layer, which consists of the $N^{(l)}$-dimensional feature embeddings for all $M$ nodes.
In the aggregate step, the (forward) message $\boldsymbol Z^{(l)} \in \mathbb  R^{M\times N^{(l-1)}}$ is computed by 
the aggregation function $\mathcal{M}^{(l)}$ that aggregates
the features $\boldsymbol H^{(l-1)}$ from the last layer 
using a modified (e.g., normalized with self-loops) adjacency matrix $\boldsymbol \Lambda \in \mathbb  R^{M\times M}$.
In the combine step, the (typically non-linear) combine function $\mathcal C^{(l)}$ 
transforms $\boldsymbol Z^{(l)}$ into the new node features $\boldsymbol H^{(l)}$ for this layer. 

A common choice for the aggregation and the combine functions in Eqs.\eqref{eq:GNN_agg} and \eqref{eq:GNN_comb}, respectively,
is 
the linear aggregation and
a one-layer perceptron,
as in a simple GCN \cite{DBLP:conf/iclr/KipfW17}:
\begin{align}
\mathcal{M}^{(l)}(  \boldsymbol{H}^{(l-1)},  \boldsymbol \Lambda) 
& = \textstyle \boldsymbol \Lambda \boldsymbol{H}^{(l-1)},
\label{eq:GCN_agg}\\
\mathcal C^{(l)}(\boldsymbol Z^{(l)})
&= \textstyle  \sigma(\boldsymbol Z^{(l)} \boldsymbol W^{(l)}),
\label{eq:GCN_comb}
\end{align}
where $\boldsymbol W^{(l)} \in \mathbb R^{N^{(l-1)} \times N^{(l)}}$ is a trainable weight matrix and $\sigma(\cdot)$ is a non-linear (entry-wise) activation.
After the final node features are computed, a readout function will be applied to make the final model predictions, e.g., sum over all nodes followed by a softmax function to produce graph-level classification probabilities.

\subsection{Explaining GNNs}

Many explanation methods for GNNs have emerged recently. These include general explanation methods, e.g., sensitive analysis (SA), guided backpropagation (GBP), class activation mapping (CAM) and excitation backpropagation (EB) \citep{DBLP:journals/corr/abs-1905-13686, DBLP:conf/cvpr/PopeKRMH19}, adapted to the GNN structure, and novel methods specialized for GNNs, e.g., GNNExplainer \citep{ying2019gnnexplainer} and PGExplainer \citep{luo2020parameterized}. GNNExplainer 
finds soft masks (for node or edge features) such that the mutual information between the predictions of the original graph and the masked graph is maximized, and uses the masks as the relevance scores. PGExplainer learns approximate discrete masks by training a parametric predictor, and masks out unimportant edges according to the learned masks. 
PGM-Explainer \citep{vu2020pgm} trains an explainable probabilistic graphical model as a surrogate of the GNN, and use its explanation as a substitution. GraphMask \citep{DBLP:conf/iclr/SchlichtkrullCT21} trains a one layered perceptron network to predict whether each edge in each layer can be removed without changing the model output significantly. XGNN \citep{DBLP:conf/kdd/YuanTHJ20} and GNNInterpreter \citep{DBLP:journals/corr/abs-2209-07924} explain a GNN on the model-level, and generates representative graphs for possible model predictions by using reinforcement learning or probabilistic generative models. GNES \citep{DBLP:conf/icdm/GaoSBYH021} is a general framework that can make the explanation more reasonable and stable by training GNN and optimizing its explanation simultaneously with specific regularization. Tage \citep{DBLP:conf/nips/XieKTH0SJ22} is a framework for efficiently explaning GNN with multiple downstream tasks, which trains a parametric explainer in a self-supervision manner.

There are also higher-order methods that consider interactions between more than two nodes. %
SubgraphX \citep{yuan2021explainability} searches the most relevant subgraph using Monte-Carlo Tree Search (MCTS) with Shapley value \citep{lund2017unified}, and applied approximation methods in computing Shapley values, which is otherwise too computation-intensive.
GNN-LRP \citep{schnake2022higher} is an LRP-based method, which scores bag-of-edges by decomposing and backpropagating the output to the input layer.
GNN-LRP considers a walk as the basic unit for attribution, which is detailed below.

\subsection{Relevance of Walks}
\label{sec:WalkRelevance}

A \emph{walk} is defined as an ordered sequence of nodes connected from layer to layer \citep{schnake2022higher}.
Assume that the whole graph $\mathcal G$
consists of $M$ nodes.  Then, a walk can be denoted by $\bfm \in \mathbb{M} $ with $\mathbb{M}  = \{1, \ldots, M\}^{L+1}$, meaning that the walk starts from the $m_0$-th node at the input layer, goes through the $m_l$-th node at the $l$-th layer, and reaches the $m_L$-th node in the last layer.
 We also denote a partial walk by $m_{l:l'}$ for $0 \leq l \leq l'  \leq L$.
   
The GNN-LRP rule 
for the MPNNs, Eqs.\eqref{eq:GNN_agg} and \eqref{eq:GNN_comb},
with the aggregation and combine functions, Eqs.\eqref{eq:GCN_agg} and \eqref{eq:GCN_comb}, 
and the ReLU activation 
is given as
\begin{align}
  \breve{  \boldsymbol  r}^{(l, m_l)} = \boldsymbol T^{l, m_l, m_{l+1}}   \breve{  \boldsymbol  r}^{(l+1, m_{l+1})},
    \label{eq:GCN_GNN_LRP_rule}
    \end{align}
    where 
    $\breve{\boldsymbol{r}}^{(l, m_l)} \in \mathbb R^{N^{(l)}}$ is the \emph{propagated relevance}
    at the node $m_l$ in the $l$-th layer,
    and $\boldsymbol T^{l, m, m'} \in \mathbb R^{N^{(l)} \times N^{(l+1)}}$ is 
    the propagation matrix whose entries are given as
\begin{align}
 T_{n, n'}^{l, m, m'} = \textstyle
\frac{\Lambda_{m, m'} H_{m, n}^{(l)} W^{\uparrow(l+1)}_{n, n'} }{\sum_{m'' ,n''} {\Lambda_{m'', m'} H_{m'', n''}^{(l)} W^{\uparrow(l+1)}_{n'', n' } }}.
\label{eq:TforLRPGamma}
\end{align}
Here
$ \boldsymbol{W}^{\uparrow}$ is a modified weight parameter depending on the choice of LRP rules \cite{bach2015pixel,DBLP:series/lncs/MontavonBLSM19,samek2021explaining,EberleBKMVM22}, e.g., 
$\boldsymbol{W}^{\uparrow} := \boldsymbol{W} + \gamma\cdot \max(0, \boldsymbol{W} )$ for the LRP-$\gamma$ rule with $\gamma \geq 0 $,
where the max operator applies entry-wise.
Note that we mostly use subscripts to specify the entry of a matrix or vector, while superscripts
for distinguishing different matrices or vectors.
For general MPNNs, \eqref{eq:GNN_agg} and \eqref{eq:GNN_comb},
relevance propagation rules \eqref{eq:GCN_GNN_LRP_rule} can be similarly defined with 
appropriate propagation matrices $\{\boldsymbol T^{l, m_l, m_{l+1}}\}$ (see an exemplary study in \citet{schnake2022higher}).
We stress that our theory and algorithms can be applied to any  GNN (possibly beyond MPNNs) as long as the propagation rule in the form of Eq.\eqref{eq:GCN_GNN_LRP_rule} is defined.

\begin{figure}[t]
\centering
\includegraphics[width=.9\linewidth]{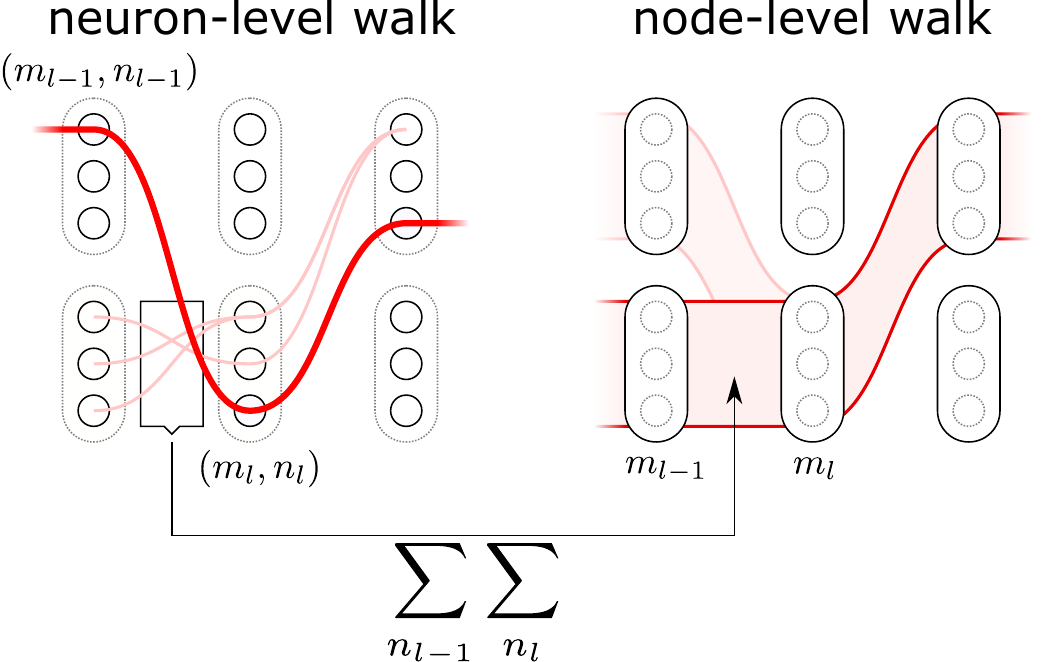}
\vspace{-3mm}
\caption{
Illustration of neuron-level and node-level walks.
The node-level path of the top neuron-level walk may differ from the top node-level walk, because many weakly relevant neuron-level walks may sum up to a strongly relevant node-level walk.
}
\label{fig:NeuronNodeLevelWalks}
\vspace{-3mm}
\end{figure}

Considering the unfolded GNN as a feed-forward neural network (FFNN),
\citet{DBLP:conf/icml/XiongSMMN22} defined the \emph{neuron-level walk} (see Fig.~\ref{fig:NeuronNodeLevelWalks}):
\begin{align}
R^{\boldsymbol{m}, \boldsymbol{n}} =\textstyle  \left(\prod_{l=0}^{L-1} T_{n_{l}, n_{l+1}}^{l, m_l, m_{l+1}} \right) 
r_{n_L}^{L, m_L},
\label{eq:NeuronlevelWalkRelevance}
\end{align}
and pointed out that it has the same decomposability as the joint distribution of a Markov chain process.
Here 
$\bfn \in \mathbb{N} \equiv \{1, \ldots, N^{(0)}\} \times \cdots \times \{1, \ldots, N^{(L)}\} $ specifies the trajectory of relevance through neurons, and
$\bfr^{L, m_{L}} \in \mathbb{R}^{N^{(L)}}$ 
denotes the neuron-level relevance at the $L$-th layer.
Based on this observation, it was shown that the propagation rule \eqref{eq:GCN_GNN_LRP_rule} for computing the (node-level) walk relevance by GNN-LRP, i.e.,
\begin{align}
R^{\boldsymbol{m}} 
 & =\textstyle \left(\prod_{l=0}^{L-1} \boldsymbol{T}^{l, m_l, m_{l+1}} \right) 
 \boldsymbol{r}^{L, m_L} 
\notag\\
& =\textstyle
\sum_{\boldsymbol{n} \in \mathbb{N}}
\left(\prod_{l=0}^{L-1} T_{n_{l}, n_{l+1}}^{l, m_l, m_{l+1}} \right) 
r_{n_L}^{L, m_L},
\label{eq:NodelevelWalkRelevance}
\end{align}
(as well as the one for the standard LRP  for general FFNN with any propagation matrices $\bfT$)
can be interpreted as the \emph{sum-product} message passing algorithm---a common tool to compute marginal probabilities of tree-structured probabilisitic graphical models \citep{DBLP:conf/aaai/Pearl82,bishop2006patternchp8}---for marginalizing over the neurons within the node-level walk $\bfm$.
\citet{DBLP:conf/icml/XiongSMMN22} applied the same argument for the subgraph relevance,
proposed in \citet{schnake2022higher},
and 
drastically accelerated the exponential computation of the original GNN-LRP to a polynomial one.

\section{Message Passing for Relevant Walk Search}
\label{sec:max-prod-gnn-lrp}

This paper aims to provide a family of efficient algorithms for finding most relevant walks; see   Section~\ref{sec:experiments} for applications. %
We reiterate from Section~\ref{sec:Introduction} that  
the same decomposability of the walk relevance as a Markov chain process (see \citet{DBLP:conf/icml/XiongSMMN22})  allows us now to propose  the application of 
 the max-product algorithm for maximization \citep{Viterbi67,bishop2006patternchp8},  instead of the sum-product algorithm for marginalization.
Specifically, we can now 
suggest two novel algorithms, namely, for neuron-level  and node-level search of the most relevant walks.

\subsection{Exact Neuron-level Search}
\label{sec:top-k-neuron-walk}

Unlike the sum-product decomposition, the max-product decomposition holds only when each factor is non-negative.  Therefore, we first maximize the absolute relevance:
\begin{align}
\breve{R}^{\boldsymbol{m},\boldsymbol{n}} =\textstyle  \left(\prod_{l=0}^{L-1} |T_{n_{l}, n_{l+1}}^{l, m_l, m_{l+1}} |\right) 
|r_{n_L}^{L, m_L} |.
\label{eq:NeuronlevelWalkAbsoluteRelevance}
\end{align}
Thanks to the decomposability of the objective,
the  maximization \eqref{eq:NeuronlevelWalkAbsoluteRelevance} can be performed by the following max-product message passing for $l = L, \ldots, 0$:
\begin{align}
\widehat{m}_l,  \widehat{n}_l
&= \textstyle
\argmax_{{m_l, n_l}} |T^{l-1, {m_{l-1}}, {m_{l}}}_{{n_{l-1}}, {n_{l}}}| \; \widehat{\mu}^{l, {m_{l}}}_{{n_l}},
\label{eq:MaxProductNeuronLevelWalk}\\
\widehat{\mu}^{l-1, {m_{l-1}}}_{{n_{l-1}}} 
&=| T_{{n_{l-1}}, {\widehat{n}_{l}}}^{l-1,m_{l-1}, \widehat{m}_l} | \; \widehat{\mu}_{\widehat{n}_l}^{l, \widehat{m}_l},
\label{eq:MaxProductNeuronLevelMessage}
\end{align}
where
$\widehat{\bfmu}^{l, m_l} \in \mathbb{R}^{N^{(l)}}$ is a (backpropagating) message at the $l$-th layer, and initialized as $\widehat{\mu}_{n_L}^{L, m_L} = |r_{n_L}^{L, m_L}|$.
 $(\widehat{m}_l, \widehat{n}_l) \in \{1, \ldots, M\}\times \{1, \ldots, N^{(l)}\}$
are also messages, and actually functions 
of $(m_{l-1}, n_{l-1})$%
---we abbreviated them  as $\widehat{m}_l = \widehat{m}_l(m_{l-1}, n_{l-1}) $ and $\widehat{n}_l = \widehat{n}_l (m_{l-1}, n_{l-1}) $
in Eqs.\eqref{eq:MaxProductNeuronLevelWalk} and \eqref{eq:MaxProductNeuronLevelMessage} to avoid from cluttering.
$(\widehat{m}_l, \widehat{n}_l)$ together specify the most relevant backward neuron-level step for any possible next step choice for the $(l-1)$-th layer.
The derivation of the message passing, \eqref{eq:MaxProductNeuronLevelWalk} and \eqref{eq:MaxProductNeuronLevelMessage}, and the detailed algorithm are given in Appendix~\ref{sec:DerivationNeuronLevelAbsoluteMaxProduct}.

The max-product message passing above finds the  neuron-level walk that has the highest absolute relevance in $\mcO(M^2 \overline{N}^2 L)$ time, where $\overline{N} = \max_{l} N^{(l)}$.  However, the solution can be not the maximizer but the minimizer of the original walk relevance \eqref{eq:NeuronlevelWalkRelevance} if it is negative.
Furthermore, we expect that only focusing on the single most relevant walk is not always informative enough for GNN explanation.  To remedy both issues, we propose a procedure to find the top-$\widetilde{K}$ absolute relevant walks, following the strategy proposed in \citet{nilsson1998efficient}.

After finding the neuron-level walk $\widehat{\bfm}^1, \widehat{\bfn}^1$ with the highest absolute relevance, we split the rest of the search space $(\mathbb{M} \times \mathbb{N}) \setminus (\widehat{\bfm}^1, \widehat{\bfn}^1)$ into $L+1$ disjoint subsets $\{\mcA_i\}$, where 
\begin{align}
   \mcA_0 &=   \{(\bfm, \bfn) : (m_0, n_0) \ne (\widehat{m}_0^{1}, \widehat{n}_0^{1})\}, \notag \\
   \mcA_i &= \{(\bfm, \bfn) : (m_{0:i-1}, n_{0:i-1}) = (\widehat{m}_{0:i-1}^{1}, \widehat{n}_{0:i-1}^{1}),
   \notag\\
   &\qquad \quad
   (m_i, n_i) \ne (\widehat{m}_i^{1}, \widehat{n}_i^{1})\}
   \mbox{ for } i = 1, \ldots, L.
    \label{eq:top-2-neuron-walk}
\end{align}
Namely, the subset $\mcA_i$ consists of all walks that have the same partial walk as the first solution until the $(i-1)$-th layer, and differ at the $i$-th layer.  No restriction is imposed for the subsequent layers, $l = i+1, \ldots, L$.
Then we apply the max-product message passing to each subset, taking the corresponding constraint into account.  Among the $(L+1)$ solutions from the subsets, the walk that gives the highest absolute relevance is the second best solution $\widehat{\bfm}^2, \widehat{\bfn}^2$. 
The third best solution can be similarly obtained by further splitting $\mcA_{\widehat{i}} \setminus (\widehat{\bfm}^2, \widehat{\bfn}^2)$, where $\mcA_{\widehat{i}}$ is the set from which the second solution was found, into disjoint subsets, and 
we can iterate this procedure 
until the top-$\widetilde{K}$ walks are collected (see Appendix~\ref{sec:DerivationNeuronLevelAbsoluteMaxProduct} for detailed procedure).
The number of disjoint subsets increases up to $\widetilde{K}L + 1$ in the worst case.
Note that this procedure can be terminated at any $\widetilde{K}$, e.g., when we are satisfied with the collection,
and therefore it is not necessary to fix $\widetilde{K}$ beforehand.

We call the algorithm described above---top-$K$ relevant neuron-level walk search by max-product message passing, \eqref{eq:MaxProductNeuronLevelWalk} and \eqref{eq:MaxProductNeuronLevelMessage},
and the search space splitting \eqref{eq:top-2-neuron-walk}---exact max-product search for neuron-level walks (EMP-neu), for which the following theorem holds (for proof see  Appendix~\ref{sec:DerivationNeuronLevelAbsoluteMaxProduct}):
\begin{theorem}
\label{thrm:ExactNeuronLevelSearch}
Assume that the top-$\widetilde{K}$ walks with the highest absolute relevance contain $K$ walks with positive relevance.  Then, 
EMP-neu finds
the top-$K$ neuron-level walks that maximize Eq.\eqref{eq:NeuronlevelWalkRelevance}
with the computational complexity 
$\mcO(LM^2 \overline{N}^2  + \widetilde{K} L^2M \overline{N})$  
and the memory cost $\mcO(LM^2 \overline{N}^2)$.
\end{theorem}
This theorem guarantees that EMP-neu can perform exact search in polynomial-time.
Notably, 
the complexity for $\widetilde{K} \geq 2$ can be much smaller than 
a naive application of the whole message passing to each subset, which would cost $\mcO(\widetilde{K} L^2M^2 \overline{N}^2) $ 
(note that the graph size $M$ and the feature dimension $\overline{N}$ are typically much larger than the network depth $L$ in large-scale problems).
This is because most of the messages required to find the $k$-th solution candidate in each subset have been already computed after the top-$(k-1)$ solutions were found, and can be reused. 
How large $\widetilde{K}$ is required for a target $K$ is affected by how many and large negative entries the propagation matrices contain.  
We empirically found (see Appendix~\ref{sec:NecessaryKTilde}) that the proportion of positive walks among the top-$\widetilde{K}$ absolute walks are more than half for LRP-$\gamma$  with  $\gamma \geq 0.2$,
and therefore typically $\widetilde{K} <  2K$.

\subsection{Approximate Node-level Search}
\label{sec:top-k-node-walk}

Let us move our focus on 
the node-level walk relevance
\eqref{eq:NodelevelWalkRelevance},
which has shown to be useful for GNN explanation \citep{schnake2022higher,DBLP:conf/icml/XiongSMMN22}.
Unfortunately, the max-product algorithm is not directly applicable to the node-level relevance because the marginalization over the neurons prevents max-product decompositions.
As a remedy, we propose an approximation method.

Let us have a close look into the maximization problem of the node-level relevance
\begin{align}
\max_{\bfm \in \mathbb{M}} R^{\boldsymbol{m}} 
& =\textstyle
\max_{\bfm \in \mathbb{M}} 
\sum_{\boldsymbol{n} \in \mathbb{N}}
\left(\prod_{l=0}^{L-1} T_{n_{l}, n_{l+1}}^{l, m_l, m_{l+1}} \right) 
r_{n_L}^{L, m_L}
\notag\\
& \hspace{-12mm}
= \textstyle
\max_{\bfm \in \mathbb{M}} 
\sum_{n_0, n_1} 
 T_{n_{0}, n_{1}}^{0, m_0, m_{1}} 
\cdots
 \sum_{n_{l}}
 T_{n_{l-1}, n_{l}}^{l-1, m_{l-1}, m_{l}} 
 \notag\\
& \hspace{-12mm} 
\textstyle
\sum_{n_{l+1}}
 T_{n_{l}, n_{l+1}}^{l, m_l, m_{l+1}}
 \cdots
 \sum_{n_{L}}
  T_{n_{L-1}, n_{L}}^{L-1, m_{L-1}, m_{L}} 
r_{n_L}^{L, m_L}.
\label{eq:AMP-aveDecomposition.First}
\end{align}
In the last equation, the summation operators are distributed according to 
the sum-product decomposition, while the maximization operator cannot be distributed.  This is because the maximization for $m_l$ involves multiple terms that depend on the propagation matrices $\{\bfT^{l', m_{l'}, m_{l'+1}}\}$ for $l' < l$ through the dependence on the neurons.

Now assume that the propagation matrices $\{\bfT^{l-1, m_{l-1}, m_{l}}\}$ at the $(l-1)$-th layer has similar columns, and can be approximated as
$\bfT^{l-1, m_{l-1}, m_{l}} \approx \overline{\bfT}^{l-1, m_{l-1}, m_{l}}$,
where $\overline{\bfT}^{l-1, m_{l-1}, m_{l}} \in \mathbb{R}^{N^{(l-1)} \times N^{(l)} }$ consists of the same columns equal to the average column of $\bfT^{l-1, m_{l-1}, m_{l}}$, i.e.,
$\overline{T}_{n_{l-1}, n_{l}}^{l-1, m_{l-1}, m_{l}} = \frac{1}{N^{(l)}} \sum_{n'_l = 1}^{N^{(l)}} T_{n_{l-1}, n'_{l}}^{l-1, m_{l-1}, m_{l}}, \forall n_{l} = 1, \ldots, N^{(l)}$. 
With this approximation, we see that the max operator can be distributed as
\begin{align}
\max_{\bfm \in \mathbb{M}} R^{\boldsymbol{m}} 
& \approx
\max_{m_0, \ldots, m_{l}} 
\sum_{n_0, n_1}
 T_{n_{0}, n_{1}}^{0, m_0, m_{1}} 
\cdots
 \overline{T}_{n_{l-1}, n_{l}}^{l-1, m_{l-1}, m_{l}} 
 \notag\\
& \hspace{-15mm} 
\max_{m_{l+1}, \ldots, m_{L}} 
 \sum_{n_{l},n_{l+1}}
 T_{n_{l}, n_{l+1}}^{l, m_l, m_{l+1}}
 \cdots
 \sum_{n_{L}}
  T_{n_{L-1}, n_{L}}^{L-1, m_{L-1}, m_{L}} 
r_{n_L}^{L, m_L}.
\notag
\end{align}

Inspired by this observation, we propose an approximate max-product search 
by averaging (AMP-ave), 
which iterates the following message passing for $l = L, \ldots, 0$:
\begin{align}
&  \widehat{m}_l =\textstyle
\argmax_{m_l}  \sum_{n_{l-1}, n_{l}} T^{l-1, {m_{l-1}}, {m_{l}}}_{{n_{l-1}}, {n_{l}}} \widehat{\mu}^{l, {m_{l}}}_{{n_l}},
\label{eq:MaxProductNodeLevelWalk}\\
& \widehat{\mu}^{l-1, {m_{l-1}}}_{{n_{l-1}}} 
= \textstyle \sum_{n_{l}}  T_{{n_{l-1}}, {{n}_{l}}}^{l-1,m_{l-1}, \widehat{m}_l} \widehat{\mu}_{{n}_l}^{l, \widehat{m}_l}.
\label{eq:MaxProductNodeLevelMessage}
\end{align}

To find the top-$K$ walks, we can apply the same search space splitting procedure \eqref{eq:top-2-neuron-walk} as in the neuron-level walk search.
Algorithm details of AMP-ave are given in Appendix~\ref{sec:DerivationNodeLevelApproximateMaxProduct}.

Below we give a few notes on AMP-ave:
\begin{itemize}

\item
Approximation error occurs in the maximization step \eqref{eq:MaxProductNodeLevelWalk}, where the marginalization over $n_{l-1}$ is already performed, ignoring the dependence of the propagation matrix $T_{{n_{l-2}}, {n_{l-1}}}^{l-2, {m_{l-2}}, {m_{l-1}}}$ at the $(l-2)$-th layer on $n_{l-1}$.  This is justified when
$(\bfT^{l-2, m_{l-2}, m_{l-1}} - \overline{\bfT}^{l-2, m_{l-2}, m_{l-1}})$ is not so large that it changes the choice $\widehat{m}_l$ of node---which we refer to the \emph{column-similarity} assumption in the subsequent sections.
We will investigate the approximation accuracy and its dependence on the LRP parameters in Section~\ref{sec:experiments}.

\item
The relevance message \eqref{eq:MaxProductNodeLevelMessage} treats the dependence on $n_{l-1}$ correctly, and therefore, the approximation error does not acumulate as long as the approximate maximization step \eqref{eq:MaxProductNodeLevelWalk} gives the correct choice.

\item 
We do not apply the absolute value operation to the propagation matrices $\{\bfT^{l, m_{l}, m_{l+1}}\}$
in the message passing.  This is because marginalizing over the absolute values tends to significantly differ from the marginalizing over the original values, and the objective in Eq.\eqref{eq:MaxProductNodeLevelWalk}  tends to be  non-negative after the neuron marginalization (see Appendix~\ref{app:max-obj-positiv} for empirical investigation).  
When negative values are involved in the maximization step, 
a few of the top-$\widetilde{K}$ walks found by AMP-ave can  have negative relevance. 
In such cases, 
we increment $\widetilde{K} (\geq K)$ until $K$ walks with positive relevance are found, 
similarly to EMP-neu.

\item
AMP-ave has the same computational complexity as EMP-neu, i.e.,
$\mcO(LM^2 \overline{N}^2  +  \widetilde{K}L^2M \overline{N})$.

\item
A naive implementation that directly works with the propagation matrices requires $\mcO(L M^2 \overline{N}^2)$ memory cost, which can be too huge for large scale problems.
Although this is inevitable for the neuron-level walk search, AMP-ave can work on the activation matrices $\{\bfH^{(l)}\}$ and the modified weight matrices $\{\bfW^{\uparrow(l)}\}$ (see Sections~\ref{sec:GraphNeuralNetworks} and \ref{sec:WalkRelevance} for the definitions), which reduces the memory costs to $\mcO(L \max(M, \overline{N})^2)$ and makes the node-level walk search feasible for larger scale problems (see Appendix~\ref{sec:deri-no-T}). For one layer in the GIN used in our experiment on Infection dataset, the memory cost reduced from $1000^2\times32^2 \approx 10^9$ to only $1000^2 + 1000*32 + 32^2 \approx 10^6$ \texttt{Tensor.float}s.

\end{itemize}

\section{Experiments}
\label{sec:experiments}

Here, we will
empirically study our proposed approach.
After introducing the datasets used in our experiments,
we first evaluate the approximation accuracy of AMP-ave, i.e., how accurately AMP-ave can find the true top-$K$ walks, by comparing with the ground-truth most relevant walks found by exhaustive search.
Then, we show qualitative results demonstrating the utility of our approach, followed by quantitative evaluations supporting the exquisite performance of our walk search approach. 
Lastly,
we report its computation cost, showing dramatic improvement over  previous GNN-LRP implementation from \citet{schnake2022higher}.
Empirical evaluation of EMP-neu in terms of accuracy (or correctness) and computation time can be found in Appendix~\ref{app:emp-neu-verify}.

\subsection{Datasets}
We use common benchmark datasets 
including {\bf BA-2motif}, {\bf MUTAG}, {\bf Mutagenicity}, and {\bf Graph-SST2} (see  Appendix~\ref{sec:models} 
for details on data and employed GNNs).
We demonstrate the scalability of our framework, 
using the {\bf Infection} dataset, for which exhaustive evaluation of all walks is infeasible.
This dataset was originally generated by simulating a dissemination process based on the susceptible-infected (SI) model---a common process in epidemiology studies \citep{bai2017optimizing, DBLP:conf/sdm/OettershagenK0M20, ISELLA2011166}.
We first generate a random directed graph with size $M$ that expresses interactions between humans, by using the code provided by  \citet{10.1145/3447548.3467283}.
Then we randomly choose $2\%$ of the population (nodes) to be the initial infected people or carriers. We simulate $L$-time steps of the infection process: each carrier infects its neighbors with probability $\lambda$, and never cures itself.
We record the infection chains from one of the initial carriers to each carrier at time $L$, which are used as the ground truth for quantitative evaluation.
We assume that the investigator, who has no information on the data generating process, trains a GNN to predict whether each person will be infected after $L$ steps.  We generated 100 samples (scenarios) with $M = 1000, L = 4, \lambda = 0.6$, and trained a $L$-layered GCN with 80 samples and tested on the other 20 samples.  The model reached  $82.51\%$ accuracy, which is close to the prediction accuracy $83.1\%$ by the \emph{oracle}.%
\footnote{
As the oracle, we estimate the infection probability of each person, 
as well as the possible infection chains with their probabilities,
by using the complete information of the data generation process including the parameter setting (see Appendix~\ref{app:oracle} for detailed computation).
They are used as the ground truth or as the best possible predictor/detector in qualitative and quantitative evaluations.
}

\subsection{Approximation Accuracy of AMP-ave}
\label{sec:acc-top-K}

\begin{figure}[t]
\begin{center}
\centerline{\includegraphics[width=\linewidth]{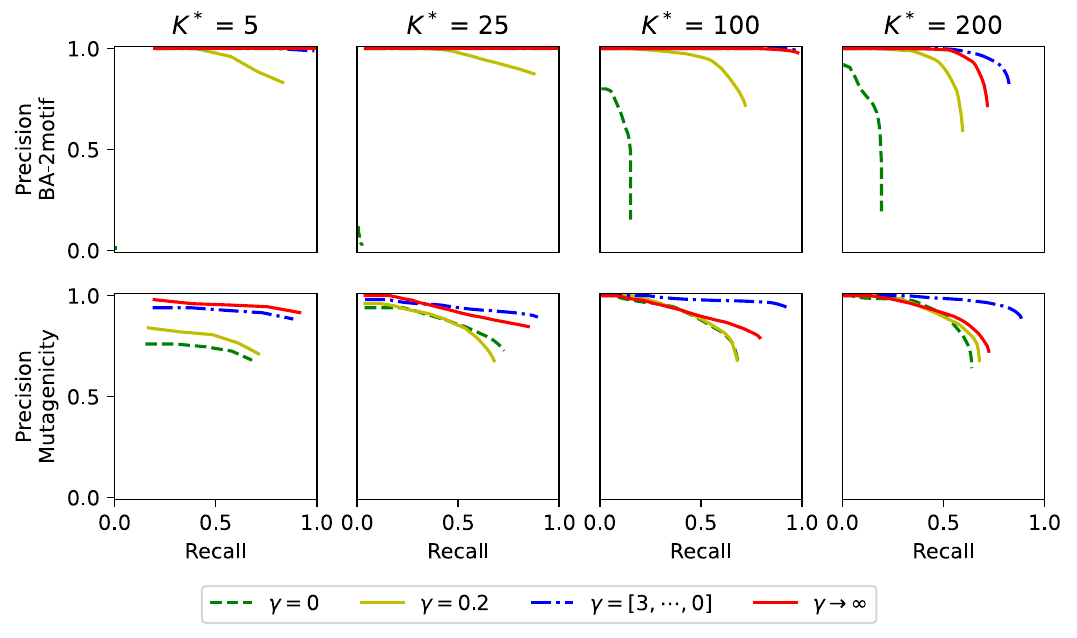}}
\vskip -0.1in
\caption{Precision-recall curves of AMP-ave in top-$K^*$ node-level walk search on BA-2motif (top) and Mutagenicity (bottom). %
}
\vskip -0.3in
\label{fig:pr-top-k}
\end{center}
\end{figure}

\begin{figure}[]
\begin{center}
\centerline{\includegraphics[width=0.8\linewidth]{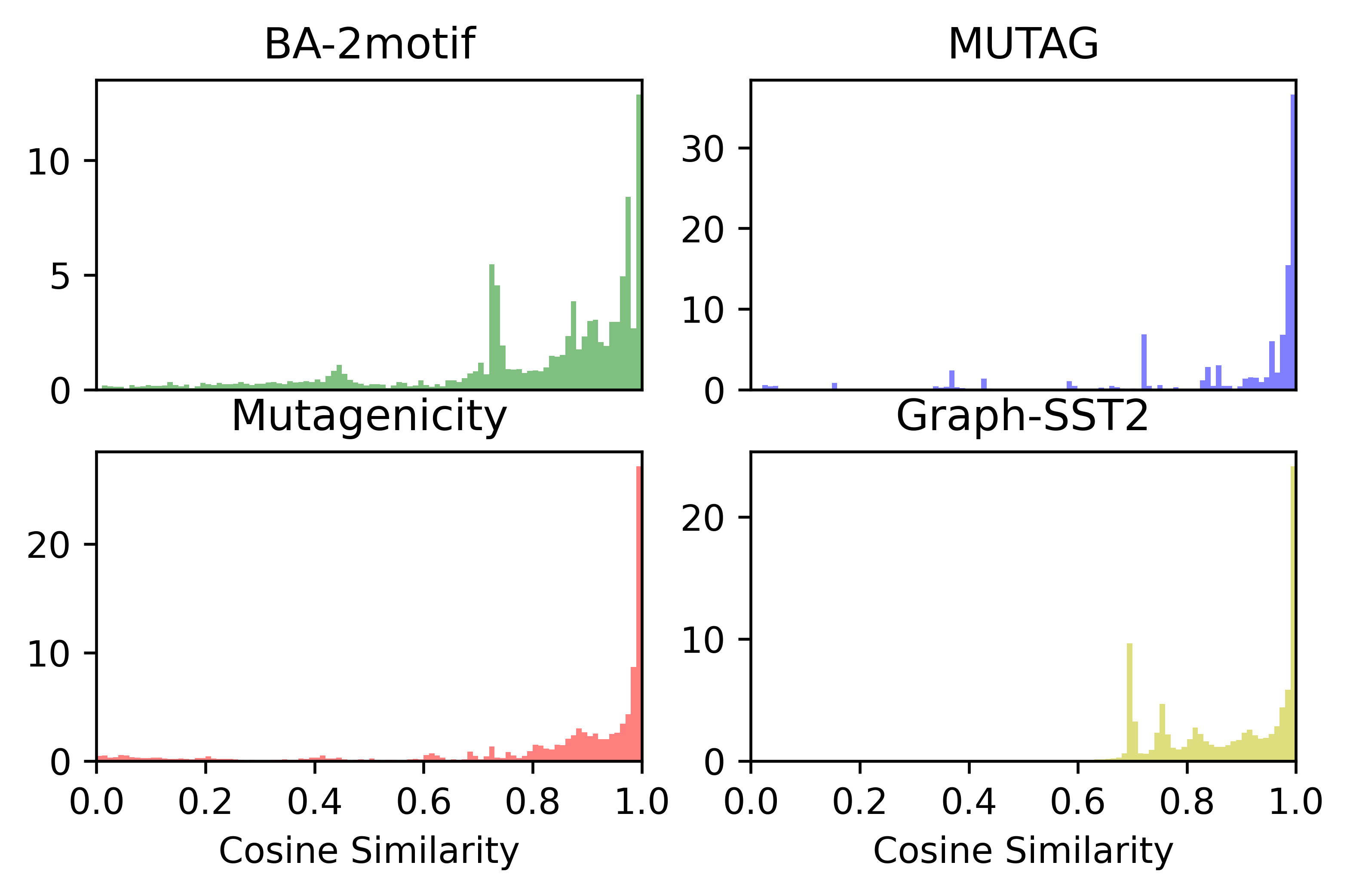}}
\vskip -0.2in
\caption{The histograms of the cosine similarity \eqref{eq:CosineSimilarity} between the column vectors and their average of the propagation matrices for $\gamma = [3, \dots, 0]$. Each panel corresponds to each dataset.
}
\label{fig:cos-sim-comb}
\end{center}
\vskip -0.3in
\end{figure}

We first validated the accuracy of AMP-ave.
By using BA-2motif, MUTAG, Mutagenicity, and Graph-SST2 datasets
with the corresponding trained GNN models, we performed an exhaustive search to identify the ground-truth top-$K^*$ walks.
Then, we performed approximate top-$K$ walk
search by AMP-ave for different $K$,
and evaluated its performance 
in terms of precision 
$\text{TP}/K$ and recall $\text{TP}/K^*$, where
$\text{TP}=|\{\text{Approx. top-$K$ walks}\} \cap \{\text{True top-$K^*$ walks}\}|$, on randomly chosen samples among the correctly classified test samples from each dataset.%
\footnote{
Due to the exponential complexity of exhaustive search,
we chose 10 samples from each dataset and excluded the Infection dataset.
}
\cref{fig:pr-top-k} shows the precision-recall curves on BA-2motif and Mutagenicity for different $K^*$ and different $\gamma$ of LRP-$\gamma$ rules. 
Here, $\gamma = [3,\cdots,0]$
indicates the recommended setting by 
\citet{schnake2022higher}, i.e., 
$\gamma$ is set from 3 to 0, linearly decreasing as $\gamma = 3(1-\frac{l}{L-1})$ for the $l$-th layer.
We observe that the approximation accuracy by AMP-ave is generally good for LRP-$\gamma$ with $\gamma \geq 0.2$.
Similar results were obtained on MUTAG and Graph-SST2 (see Appendix~\ref{app:pr-graphsst2}).
Note that the accuracy of AMP-ave is low for LRP-$0$, which however is rarely used for GNN explanation because of its poor performance in general \citep{schnake2022higher}.
In the subsequent experiments, we focus on the recommended setting $\gamma = [3,\cdots,0]$.\\
{\bf  Column-similarity Assumption}:
Here, we investigate to what extent the assumption required for high accuracy of AMP-ave holds.  Specifically, we measure the cosine similarity between the column vectors $\{{\bft}_{n_l}^{l, m_{l}, m_{l+1}}\}_{n_l=1}^{N^{(l)}}$ of the propagation matrix ${\bfT}^{l, m_{l}, m_{l+1}}$ with their average $\overline{\bft}^{l, m_{l}, m_{l+1}}$ (the corresponding column vectors of $\overline{\bfT}^{l, m_{l}, m_{l+1}}$) for each edge $(m_l, m_{l+1})$ in each layer $l$:
   \begin{align}
   \textstyle
   \frac{\left(\bft^{l, m_{l}, m_{l+1}}_{n_{l}}\right)^\T \overline{\bft}^{l, m_{l}, m_{l+1}}_{n_{l}}}{\left|\bft^{l, m_{l}, m_{l+1}}_{n_{l}}\right| \;\left|\overline{\bft}^{l, m_{l}, m_{l+1}}_{n_{l}}\right|}
   \quad \mbox{for } \quad n_l = 1, \ldots, N^{(l)}.
   \label{eq:CosineSimilarity}
   \end{align}
   \cref{fig:cos-sim-comb} shows the histograms of the cosine similarity over all $n_l, m_l, m_{l+1}, l$ and all positive class data samples classified correctly.  The zero column vectors are excluded.
   The average cosine similarity is above $0.8$ for all four datasets, which explains the good accuracy of AMP-ave. Further analysis is necessary to guarantee the approximation accuracy, and fully understand its relation to the column similarity.

\begin{figure*}[t]
\vskip -0.1in
\begin{center}
  \subfigure[AMP-ave (top-3 walks).]{\includegraphics[scale=0.5]{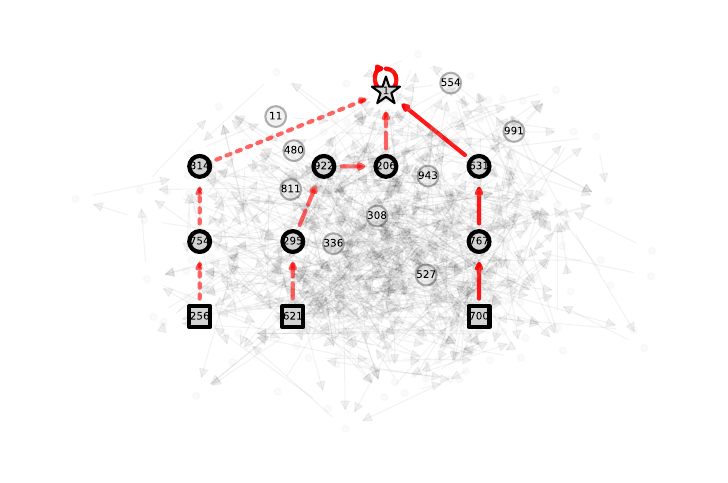}\label{fig:infection_ours}} 
  \hspace{-12mm}
    \subfigure[Node-IG (top 10 nodes).]{\includegraphics[scale=0.5]{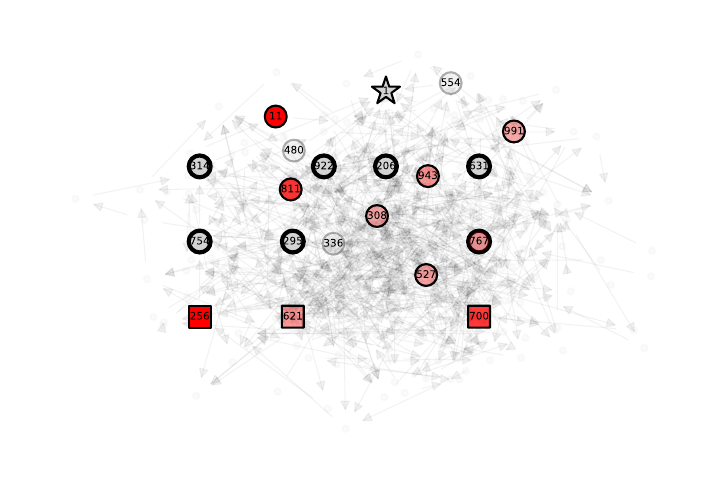}\label{fig:infection_node}} 
  \hspace{-12mm}    
  \subfigure[Edge-IG (top 10 edges).]{\includegraphics[scale=0.5]{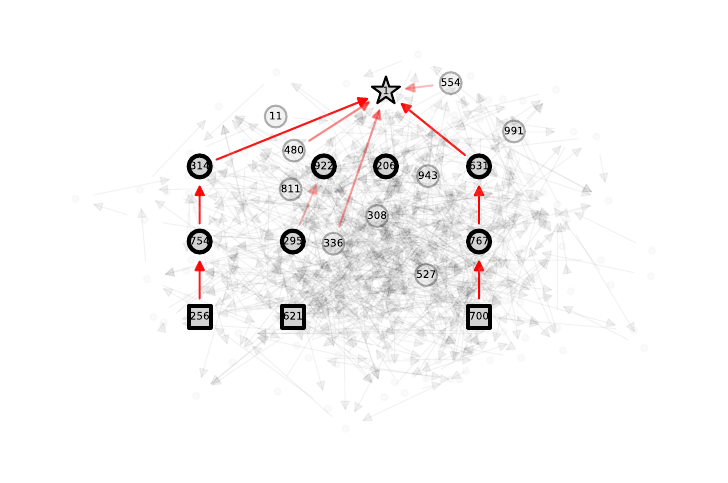}\label{fig:infection_edge}} 
\vspace{-3mm}
  \caption{ Visual explanation by AMP-ave (ours), Edge-IG, and Node-IG on Infection dataset. The deeper the red color is, the higher the relevance is. The star node at the top is the target node for which the prediction is explained, while the square nodes at the bottom are initial carriers. For clarity, we 
  only plot the nodes within $4$-hops from the target node, 
  and the nodes involved in the 3 possible infection chains are depicted as thick circles.
  The top-1 walk by AMP-ave, shown  as solid arrows, is exactly the ground truth infection chain, with the top-2 and top-3 walks being two other possible infection chains.
  }
  \vspace{-4mm}
\label{fig:infection-vis}
\end{center}
\end{figure*}

\subsection{Visualization with Top-$K$ Walks}

Now we demonstrate that AMP-ave provides better explanations than the lower-order methods on the large-scale Infection dataset, for which a naive implementation for GNN-LRP cannot be applied.
We also demonstrate
on smaller-scale Mutagenicity and Graph-SST2 datasets
that 
the top-$K$ walks, efficiently found by AMP-ave, already capture the most important information that GNN-LRP can get by evaluating all walks.\\
\begin{figure}[t]
\begin{center}
  \includegraphics[scale=0.5]{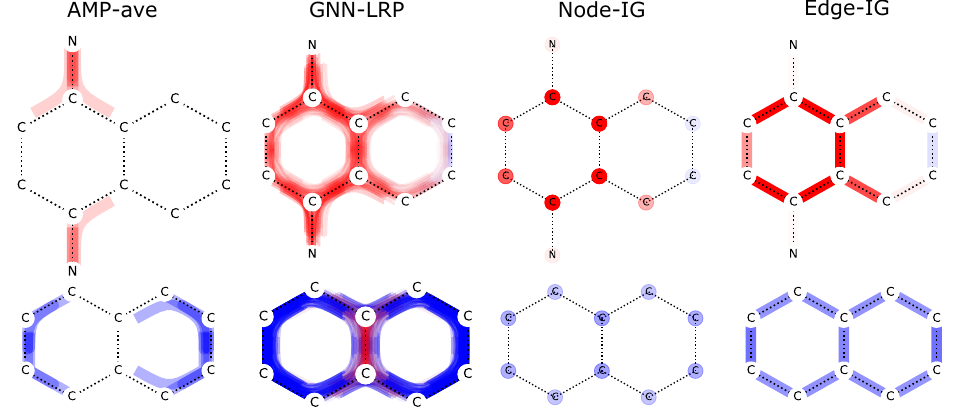}
\vspace{-3mm}
  \caption{Explanation for 1,4-Naphthalenediamine (top; mutagenic), and  
Naphthalene (bottom; non-mutagenic) from Mutagenicity. 
  The red and blue colors indicate evidence of mutagenicity and non-mutagenicity, respectively. Top-10 walks by our AMP-ave can already capture the most relevant functional groups for molecules' (non-)mutagenicity with polynomial complexity. 
  }
  \vspace{-5mm}
\label{fig:mutagenicity-vis}
\end{center}
\vskip -0.15in
\end{figure}
{\bf  Infection}: \cref{fig:infection-vis} compares explanations by our AMP-ave to Node-IG and Edge-IG \citep{DBLP:conf/icml/SundararajanTY17} on the Infection dataset with $M = 1000, L=4$.
Note that Edge-IG showed state-of-the-art performance on a similar infection chain detection task \citep{10.1145/3447548.3467283},
outperforming the other existing methods including Grad-CAM \citep{DBLP:journals/corr/abs-1905-13686}, GNNExplainer \citep{ying2019gnnexplainer} and PGMExplainer \citep{vu2020pgm}.
In the figure, each method explains why the GNN predicts that Node 1, depicted as a star at the top, will be infected after $L$ time steps.  The initial carriers that can infect the target node are depicted as squares at the bottom, and the nodes on the possible infection chains, identified by the oracle, are depicted with thick circles.
In this example, there are only 3 possible infection chains, which were identified by AMP-ave as the top-3 relevant walks.
On the contrary, Node-IG and Edge-IG do not identify those nodes as the most relevant nodes or the nodes connected to the most relevant edges.
We also observe another 
advantage of AMP-ave that users can easily identify the whole infection chains, unlike the node-/edge-level explanations that only provide 
partial information.\\
{\bf Chemistry}: \cref{fig:mutagenicity-vis} shows example explanations for the mutagenicity prediction by a GNN on 1,4-Naphthalenediamine (top) and Naphthalene (bottom) from Mutagenicity dataset.  The former is known as mutagenic, while the latter as non-mutagenic.
The figure compares our AMP-ave with $K = 10$, to Node-IG, Edge-IG, as well as to GNN-LRP that depicts all $M^{L+1}$ walks.
The red and blue colors indicate evidence for mutagenicity and non-mutagenicity, respectively.
We observe that the top-$10$ walks found by AMP-ave already capture the functional group------the $\text{NH}_2$ combined with an aromatic carbon ring---that are known to cause mutagenicity.
AMP-ave also found the non-fused carbon rings, which indicates non-mutagenicity.
GNN-LRP also provides negative evidence to the (non-)mutagenicity of the molecules, which is also useful: in the upper molecule the right carbons are indicators for non-mutagenicity, and in the bottom molecule the middle C-C points to a fused carbon ring
could in principle be an indicator for mutagenicity in some molecules.
However, we emphasize that our AMP-ave, which captures the most important positive evidence with only polynomial complexity, is a useful alternative to the full GNN-LRP, which requires exponential complexity. 
Other examples are shown in
Appendix~\ref{sec:AdditionalVisualization}.\\
{\bf  Language}: We furthermore compared AMP-ave with the baseline methods on Graph-SST2 dataset, and observed similar trends as shown for infection and chemistry above  (see Appendix~\ref{sec:AdditionalVisualization}).

\subsection{Quantitative Evaluation}

\begin{figure}[t]
\begin{center}
\centerline{\includegraphics[width=0.75\linewidth]{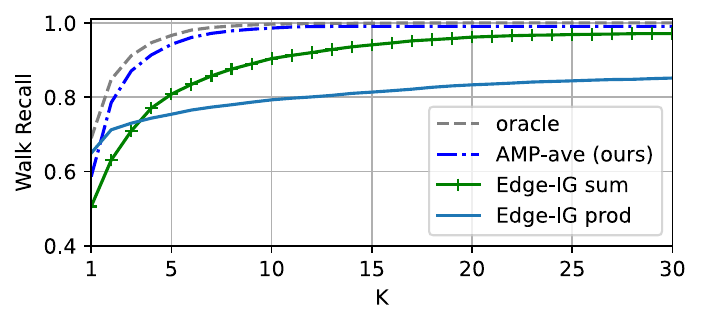}}
\vskip -0.15in
\caption{
Recall of infection chain detection on Infection dataset. 
}

\label{fig:infection-exp}
\end{center}
\vskip -0.4in
\end{figure}

\begin{figure}[t]
\begin{center}
\centerline{\includegraphics[width=0.75\linewidth]{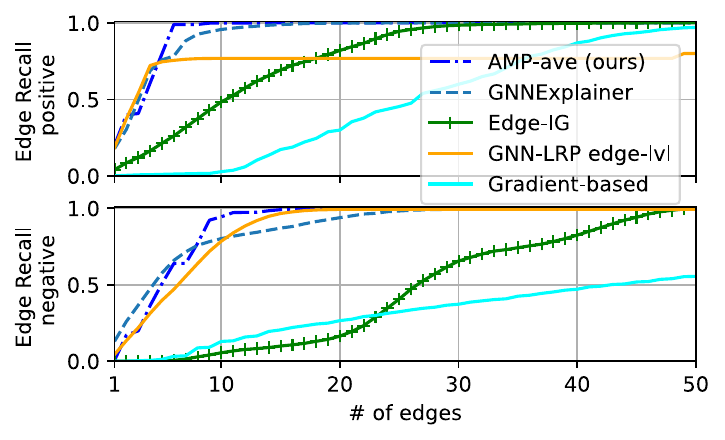}}

\vskip -0.2in
\caption{Recall of motif's edge detection on BA-2motif dataset. Positive samples and negative samples are plotted separately.}
\label{fig:ba2motif-edge-detection}
\end{center}
\vskip -0.4in
\end{figure}

We conducted a quantitative evaluation on Infection data by using the ground truth infection chain,
which well-trained GNNs are expected to use as an important information flow.
 We used AMP-ave as an infection chain detector, and evaluated its performance with different $K$.
As shown in \cref{fig:infection-exp}, top-5 walks already include the ground-truth walk with $94.22\%$ recall, and the performance is close to the oracle detector (see Appendix~\ref{app:oracle}).
Since no existing polynomial-time method provides walk-level explanation, we compare our method with two heuristic methods, Edge-IG sum and Edge-IG prod, based on Edge-IG \citep{DBLP:conf/icml/SundararajanTY17}, where top $K$ walks are constructed from the most relevant edges (see Appendix~\ref{app:edge-ig-walk} for details).
\cref{fig:infection-exp} shows that the two heuristics are clearly outperformed by our AMP-ave.\\
We also used the BA-2motif dataset, which provides the ground truth subgraphs as motifs,
and evaluated how accurately explanation methods can detect the motifs.
Specifically, we labeled all edges included in the motif as positive samples,
and compared edge detection performance of AMP-ave
to popular edge-level explanability baseline methods with comparable computational complexity,%
\footnote{We excluded in our evaluation recently proposed methods, e.g.,  GraphMask \citep{DBLP:conf/iclr/SchlichtkrullCT21} and Tage \citep{DBLP:conf/nips/XieKTH0SJ22}, that are by several orders of magnitude slower than ours.}
including Edge-IG, GNNExplainer, edge-level GNN-LRP (relevance propagated to edges in the input layer) and simple Gradient-based heatmap for edges.
Here the edge scoring by AMP-ave is simply the highest relevance of the walk that contains the corresponding edge.
 \cref{fig:ba2motif-edge-detection} shows the recall.
We see that AMP-ave reaches 100\% recall faster than all baselines, indicating that it can detect the motif more precisely.

\subsection{Computational Efficiency}

\begin{figure}[t]
 \vskip -0.0in
\begin{center}
  \subfigure[
  Network depth dependence 
]{\includegraphics[scale=0.4]{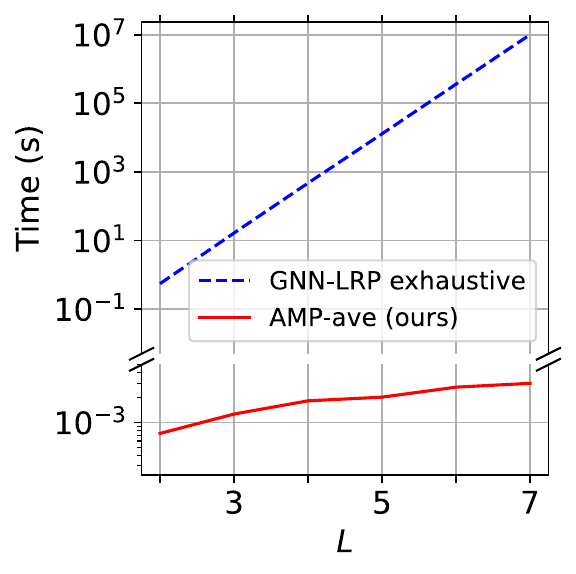}\label{fig:time-L}} 
  \subfigure[
  Graph size dependence 
  ]{\includegraphics[scale=0.4]{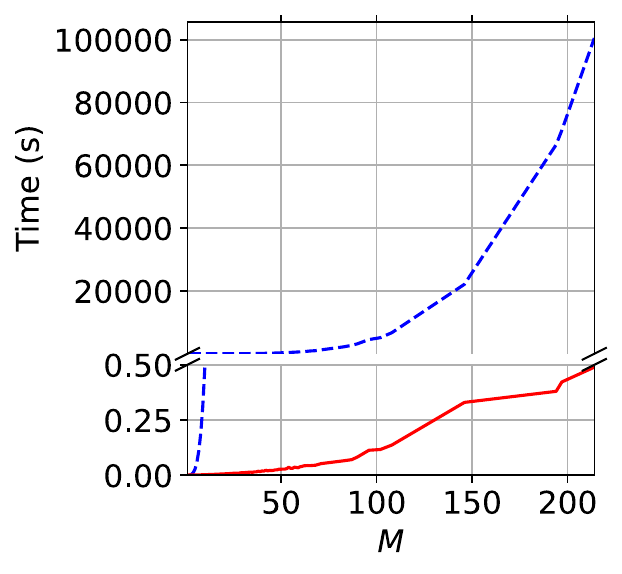}\label{fig:time-M}} 
\vspace{-4mm}
  \caption{Computation time dependence on the network depth $L$ (left) and on the graph size $M$ (right). Note the different vertical scales in the top and bottom parts,
  and that y-axis in (a) is in log-scale. 
For the depth larger than $L > 4$, the computation time of  exhaustive search is estimated from partial computation, since the whole computation is infeasible.
(a) Top-1 walk search with GIN-$L$ for $L = 2, \ldots, 7$  
  on BA-2motif.
   (b) Top-1 walk search with GIN-$3$ for $M$s among all samples in Mutagenicity dataset.   
  }
  \vspace{-5mm}
\label{fig:runtime_baselines}
\end{center}
\end{figure}

\cref{tab:time-ba2motif}
shows computation time (on an M1Pro CPU) of explanation methods on the BA-2motif and Infection datasets.  AMP-ave is orders of magnitude faster than GNN-LRP, where all walks are evaluated,
and can be applied {\em even} to the large Infection dataset.
The computation time of Edge-IG and GNNExplaner is measured in the task of relevant edge detection.
\begin{table} [t]
 \vskip -0.1in
    \caption{Computation time (in seconds). 
    B and I denote BA-2motif (small graph) and Infection (large graph) datasets, respectively. 
    }
    \label{tab:time-ba2motif}
    \vskip 0.1in
    \begin{center}
        \begin{small}
            \begin{sc}
                \begin{tabular}{lrr}
                    \toprule
                                  & Time (B) & Time (I) \\
                    \midrule

                    AMP-ave $K = 1$ & 0.003 & 0.137\\
                    AMP-ave $K = 25$ & 0.123 & 1.217\\
                    AMP-ave $K = 1000$ & 2.574 & 86.006\\
                    Edge-IG (edge-level) & 0.124 & 0.514\\
                    GNNExplainer (edge-level) & 0.371 & 60.621\\
                    GNN-LRP (exhaustive) & 15.879 & $>10^{11}$\\
                    \bottomrule
                \end{tabular}
            \end{sc}
        \end{small}
    \end{center}
    \vskip -0.2in
\end{table}
\cref{fig:runtime_baselines}
plots the computation time of AMP-ave and exhaustive search as functions of the network depth $L$ (left) and the graph size $M$ (right). The huge computational gain by our approach becomes evident.

\section{Conclusion}

Many practical applications such as chemistry, infection biology, NLP etc., embody complex correlations that are higher-order in nature. While deep learning models can exploit such structure for improving prediction, it has so far been a challenge to practically extract such higher-order information from a learned model.\\ %
The notion of walk relevance, based on layer-wise relevance propagation for graph neural networks (GNN-LRP) \citep{schnake2022higher}, was introduced for higher-order attribution; notably it has also provided a new perspective of LRP computation as a byproduct---the decomposability of relevance.
Previous work used this property and developed efficient marginalization algorithms, called subgraph GNN-LRP, for subgraph attribution
by using the sum-product message passing \citep{DBLP:conf/icml/XiongSMMN22},
while in this paper we propose a family of novel relevant walk search algorithms based on max-product message passing.
Specifically, our proposed EMP-neu performs exact search for the top-$K$ most relevant walks at neuron-level, and
AMP-ave performs 
approximate search at node-level, both in {\em polynomial-time} instead of previous {\em exponential time} approaches (see \citet{schnake2022higher}). 
Our novel methods provide robust and fast explanation with their performance  validated competitively in our experiments on several synthetic and real-world datasets.
In future studies we will aim to apply our novel efficient higher-order explanation methods broadly to the sciences and engineering, e.g., epidemic studies \citep{DBLP:conf/sdm/OettershagenK0M20, ISELLA2011166} and software vulnerability detection \citep{DBLP:conf/sp/YamaguchiGAR14}, ultimately hoping to contribute to furthering the transparency, security, robustness, and fairness of machine learning methods.\\
Finally, we would like to stress that our novel max-product approach can be  applied to any feed-forward neural network: EMP-neu is for finding neuron-level walks, while AMP-ave is for finding block-level walks, where marginalization over the neurons in blocks is involved. An immediate application would be 
to obtain neuron-/block-level decompositions similar to \citet{Achtibat22} for discovering concept-based decompositions of even finer granularity.

\subsection*{Limitations}
Our node-level walk search method---AMP-ave---is an approximation method, of which the accuracy has not been theoretically guaranteed but only supported by empirical evaluation.  Further investigation on the relation between the accuracy of AMP-ave and properties of propagation matrices is necessary to understand for what propagation rules AMP-ave is reliable.  A compromising approach is also possible and worth pursuing: one can cluster the neurons based on the propagation vectors so that the column-similarity assumption better holds at the expense of the computational cost proportional to $N^L$, where $N$ is the number of clusters.\\
Another limitation is that our approach can be applied only to the models for which the relevance propagation can be defined as in Eq.\eqref{eq:GCN_GNN_LRP_rule}.  This might exclude some of the general GNNs beyond MPNNs.   Efforts should be made for developing appropriate propagation rules for different architectures, in order to explain general large scale GNNs with our efficient algorithms.

\section*{Acknowledgments}

This work was supported by the German Ministry for Education and Research (BMBF) as BIFOLD - Berlin Institute for the Foundations of Learning and Data under grants 01IS18025A and  01IS18037A.

\bibliography{main}
\bibliographystyle{icml2023}

\newpage
\appendix
\onecolumn

\section{Notation Table}
\label{sec:notation_table}

Table~\ref{tab:notation_table} summarizes the notation used in this paper.

\begin{table}[h]
    \caption{Notation.}
    \label{tab:notation_table}
    \begin{center}
        \begin{small}
                \begin{tabular}{|c||p{0.31\textwidth}|}
                    \hline
                    $h, \boldsymbol h, \boldsymbol H$, $H_{m,m'}$ & scalar, vector, matrix, matrix entry \\
                    $ m_{l:l'} $ & partial vector with indices $(l, \dots, l') $ \\ %
                    \hline
                    $\mathcal G$ and $ \mathcal S$ & graph and subgraph \\
                    $\boldsymbol m$ and $ \boldsymbol n$ & sequence of nodes and neurons \\
                    $m$, $m_l$ & integers for node identifications \\
                    $n$, $n_l$ & integers for neuron identifications \\
                    $R$, $\boldsymbol r$ & relevance \\
                    $\breve{\boldsymbol r}$ & propagated relevance, message, or belief \\
                    $\boldsymbol T$ & propagation matrix\\                   
                    $\widehat R, \widehat \bfm, \widehat \bfn $ & optimum relevance, optimum walk, etc. \\
 
                    \hline
                \end{tabular}
        \end{small}
    \end{center}
\end{table}

\section{Details of Exact Max-product Message Passing Algorithm for Neuron-level Search (EMP-neu)}
\label{sec:DerivationNeuronLevelAbsoluteMaxProduct}

\subsection{EMP-neu-Basic: Derivation of Message Passing 
Equations \eqref{eq:MaxProductNeuronLevelWalk} and \eqref{eq:MaxProductNeuronLevelMessage}}
\label{sec:DerivationEmpNeuBasic}

\begin{algorithm}[b!]
   \caption{Find the highest absolute relevant neuron-level walk (EMP-neu-Basic)}
   \label{alg:most-rel-neur-walk}
\begin{algorithmic}
   \STATE {\bfseries Input:} \# of nodes: $M$, \# of neurons at  $l$-th layer: $N^{(l)}$, LRP transition matrices $\{\boldsymbol T^{l, m_l, m_{l+1}}\}$, initial messages $\{\widehat{\bfmu}^{L, m_L}\}$ such that $\widehat{\mu}_{n_{L}}^{L, m_L} = |r_{n_{L}}^{L, m_L}|$.
   
   \FOR{$l=L$ {\bfseries to} $1$}
       \FOR{$\textcolor{blue}{m_{l-1}}=1$ {\bfseries to} $M$}
           \FOR{$\textcolor{blue}{n_{l-1}}=1$ {\bfseries to} $N^{(l-1)}$}
               \STATE Find $\textcolor{red}{m_l, n_l}$ by solving: 
               
               $\argmax_{\textcolor{red}{m_l, n_l}} |T^{l-1, \textcolor{blue}{m_{l-1}}, \textcolor{red}{m_{l}}}_{\textcolor{blue}{n_{l-1}}, \textcolor{red}{n_{l}}}| \widehat{\mu}^{l, \textcolor{red}{m_{l}}}_{\textcolor{red}{n_l}}$
               \STATE Store the result in a maximum step mapping $(\textcolor{blue}{m_{l-1}, n_{l-1}}) \rightarrow (\textcolor{red}{m_{l}, n_{l}})$.
               \STATE Compute the relevance along the corresponding maximum walk step for $\textcolor{blue}{m_{l-1}, n_{l-1}}$:

               $\widehat{\mu}^{l-1, \textcolor{blue}{m_{l-1}}}_{\textcolor{blue}{n_{l-1}}} = |T^{l-1, \textcolor{blue}{m_{l-1}}, \textcolor{red}{m_{l}}}_{\textcolor{blue}{n_{l-1}}, \textcolor{red}{n_{l}}}| \widehat{\mu}^{l, \textcolor{red}{m_{l}}}_{\textcolor{red}{n_l}}$
           \ENDFOR
       \ENDFOR
   \ENDFOR

   \STATE Select $(m_0^*, n_0^*)$ by $\argmax_{m_0, n_0} \widehat{\mu}^{0, \textcolor{black}{m_{0}}}_{\textcolor{black}{n_0}}$.
   \STATE Read from the maximum step mappings the full walk $(\bfm^*, \bfn^*)=(m_0^*,n_0^*) \rightarrow (m_1^*,n_1^*) \rightarrow \cdots \rightarrow (m_L^*,n_L^*)$.
   \STATE {\bfseries return} $(\bfm^*, \bfn^*)$.
\end{algorithmic}
\end{algorithm}

Applying 
the max-product decompositions to the objective \eqref{eq:NeuronlevelWalkAbsoluteRelevance}
gives
\begin{equation}
\begin{aligned}
\widehat{R} \equiv  \max_{\bfm, \bfn} \breve{R}^{\bfm, \bfn}
&=  \max_{\bfm, \bfn}|T^{0, m_0, m_{1}}_{n_0, n_1}|
   |T^{1, m_1, m_{2}}_{n_1, n_2}|    
    \cdots 
    |T^{L-1, m_{L-1}, m_{L}}_{n_{L-1}, n_{L}}| |r^{L, m_{L}}_{n_L}| \\
    &= \max_{m_0, n_0}
    \underbrace{
    \max_{m_1, n_1}|T^{0, m_0, m_{1}}_{n_0, n_1}| 
    \underbrace{
    \max_{m_2, n_2}
     |T^{1, m_1, m_{2}}_{n_1, n_2}| \cdots     
 \underbrace{  \max_{m_L, n_L}|T^{L-1, m_{L-1}, m_{L}}_{n_{L-1}, n_{L}}|
 \underbrace{|r^{L, m_{L}}_{n_L}|  
 }_{= \widehat{\mu}_{n_{L}}^{L, {m}_{L}}}
 }_{= \widehat{\mu}_{n_{L-1}}^{L-1, {m}_{L-1}}}
 }_{= \widehat{\mu}_{n_{1}}^{1, {m}_{1}}}
}_{= \widehat{\mu}_{n_{0}}^{0, {m}_{0}}},
    \notag
\end{aligned}
\end{equation}
where $\{\widehat{\mu}_{n_{l}}^{l, {m}_{l}}\}$   are the messages that can be computed by Eqs.\eqref{eq:MaxProductNeuronLevelWalk} and \eqref{eq:MaxProductNeuronLevelMessage} sequentially for 
$l = L, \ldots, 0$.
Thus, we get $\widehat{R} = \max_{m_0, n_0} \widehat{\mu}_{n_{0}}^{0, {m}_{0}}$.
Backtracing the mappings $(m_{l-1}, n_{l-1}) \rightarrow (m_l, n_l)$, computed by Eq.\eqref{eq:MaxProductNeuronLevelWalk}, for $l = 1, \ldots, L$ gives the walk maximizing the absolute relevance.
The algorithm, which we call \textbf{EMP-neu-Basic}, for finding the highest absolute relevant neuron-level walk is summarized in \cref{alg:most-rel-neur-walk}.

\subsection{Complexity of Finding the Most Relevant Walk by EMP-neu-Basic}
\label{app:top1-complexity}
From \cref{alg:most-rel-neur-walk} we can find that the computational complexity of finding the highest absolute relevant neuron-level walk is $\mathcal O(LM^2\overline{N}^2+MN+L) = \mathcal O(LM^2\overline{N}^2)$. The memory cost including the storage for the transition matrices and the maximum step mappings is  $\mathcal O(LM^2\overline{N}^2)$.

\begin{algorithm}[t]
   \caption{Search space splitting for finding the top-$\widetilde{K}$ most relevant neuron-level walks}
   \label{alg:top-k-rel-neur-walk}
\begin{algorithmic}
   \STATE {\bfseries Input:} top-1 most relevant walk $(\bfm^1, \bfn^1)$, maximum step mappings $({m_{l-1}, n_{l-1}}) \rightarrow ({m_{l}, n_{l}})$.
   \STATE Initialize $SearchSpace = \{\}$
   \STATE Initialize $TopKWalks = \{(\bfm^1, \bfn^1)\}$
        \FOR{$l=0$ {\bfseries to} $L$}
           \STATE $subset = \{(\bfm, \bfn): (m_j, n_j) = (m_j^{1}, n_j^{1}) \ \forall{j<l}, (m_l, n_l) \ne (m_l^{1}, n_l^{1})\}$.
                \STATE Set the beginning of $(\bfm^*, \bfn^*)$ such that $(m_j^*, n_j^*) = (m_j^{1}, n_j^{1}) \ \forall j<l$.
                \STATE Read from the maximum step mappings the following steps $(m^*_l, n^*_l) \rightarrow \cdots \rightarrow (m^*_L, n^*_L)$.
                \STATE Add $(subset: (\bfm^*, \bfn^*))$ to $SearchSpace$.
        
        \ENDFOR
   \FOR{$k=2$ {\bfseries to} $\widetilde{K}$}

        \STATE $MaxRelevance = -\infty$
        \FOR{\{$subset: (\bfm^*, \bfn^*)\}$ {\bfseries in} $SearchSpace$}
            \IF{ $R^{(\bfm^*, \bfn^*)} > MaxRelevance$} 
                \STATE $MaxRelevance = R^{(\bfm^*, \bfn^*)}$.
                \STATE $(\bfm^k, \bfn^k) = (\bfm^*, \bfn^*)$
                \STATE $MaxSubset = subset$
            \ENDIF
        \ENDFOR

        \STATE Add $(\bfm^k, \bfn^k)$ to $TopKWalks$.
        \STATE Split $MaxSubset$ according to \eqref{eq:top-3-neuron-walk.app}, read out each subspace' most relevant walk, and add them to $SearchSpace$.
        \STATE Remove $MaxSubset$ from $SearchSpace$.
   \ENDFOR
   \STATE {\bfseries return} $TopKWalks$.
\end{algorithmic}
\end{algorithm}

\subsection{Search Space Splitting for Finding Top-$\widetilde{K}$ Walks}
\label{sec:App.SearchSpaceSplitting}

Let us denote by $(\widehat{\bfm}^1, \widehat{\bfn}^1)$ the best solution found by EMP-neu-Basic.
Then, the second solution can be found by splitting the rest of the search space $(\mathbb{M} \times \mathbb{N}) \setminus (\widehat{\bfm}^1, \widehat{\bfn}^1)$ into $L+1$ disjoint subsets $\{\mcA_i\}$, where 
\begin{align}
   \mcA_0 &=   \{(\bfm, \bfn) : (m_0, n_0) \ne (\widehat{m}_0^{1}, \widehat{n}_0^{1})\}, \notag \\
   \mcA_i &= \{(\bfm, \bfn) : (m_{0:i-1}, n_{0:i-1}) = (\widehat{m}_{0:i-1}^{1}, \widehat{n}_{0:i-1}^{1}),
   (m_i, n_i) \ne (\widehat{m}_i^{1}, \widehat{n}_i^{1})\}
 \quad  \mbox{ for }  \quad i = 1, \ldots, L.
    \label{eq:top-2-neuron-walk.app}
\end{align}
Note that the subset $\mcA_i$ consists of all walks that have the same partial walk as the first solution until the $(i-1)$-th layer, and differ at the $i$-th layer. No restriction is imposed for the subsequent layers, $l = i+1, \ldots, L$.
Then we apply EMP-neu-Basic to each subset, taking the corresponding constraint into account.  Among the $(L+1)$ solutions from the subsets, the walk that gives the highest absolute relevance is the second best solution $(\widehat{\bfm}^2, \widehat{\bfn}^2)$. 

Assume that the second solution was found from $\mcA_{\widehat{i}}$, which means that 
\[
(\widehat{m}_{0:i-1}^{1}, \widehat{n}_{0:i-1}^{1}) = (\widehat{m}_{0:i-1}^{2}, \widehat{n}_{0:i-1}^{2}).
\]
To obtain the third best solution, we split $\mcA_{\widehat{i}} \setminus (\widehat{\bfm}^2, \widehat{\bfn}^2)$
into $(L-\widehat{i}+1)$ disjoint 
subsets $\{\mcA_{\widehat{i}, j}\}$, where 
\begin{align}
   \mcA_{\widehat{i}, 0} &=   \{(\bfm, \bfn) \in \mcA_{\widehat{i}} : (m_{\widehat{i}}, n_{\widehat{i}}) \notin \{(\widehat{m}_{\widehat{i}}^{1}, \widehat{n}_{\widehat{i}}^{1}),
   (\widehat{m}_{\widehat{i}}^{2}, \widehat{n}_{\widehat{i}}^{2})\}
   \}, \notag \\
   \mcA_{\widehat{i}, j} &= \{(\bfm, \bfn) \in \mcA_{\widehat{i}} : (m_{\widehat{i}:\widehat{i} + j -1}, n_{\widehat{i}:\widehat{i}+j-1}) = (m_{\widehat{i}:\widehat{i} + j -1}^{2}, n_{\widehat{i}:\widehat{i} + j -1}^{2}),
   (m_{\widehat{i} + j}, n_{\widehat{i} + j}) \ne (\widehat{m}_{\widehat{i} + j}^{2}, \widehat{n}_{\widehat{i} + j}^{2})\}
   \notag\\
   &\qquad \qquad\qquad \qquad \qquad\qquad
 \quad  \mbox{ for }  \quad j = 1, \ldots, L - \widehat{i},
    \label{eq:top-3-neuron-walk.app}
\end{align}
and apply EMP-neu-Basic to each subset.
Now, we have $(2L - \widehat{i} +1)$ disjoint subsets that covers the remaining search space, i.e.,
\begin{align}
(\cup_{i \ne \widehat{i}} \mcA_i)
\cup
(
\cup_{j =0}^{L- \widehat{i}} \mcA_{\widehat{i}, j}
)
=
(\mathbb{M} \times \mathbb{N}) \setminus \{(\widehat{\bfm}^1, \widehat{\bfn}^1) \cup (\widehat{\bfm}^2, \widehat{\bfn}^2)\},
\notag
\end{align}
with the maximizer from each subset.
The third best solution is the best one among those maximizer.

Similarly, the $k$-th solution can be found by splitting the subset from which the $(k-1)$-th solution was found into disjoint subsets, and applying EMP-neu-Basic to each new subset.
We continue this process 
 until top-$\widetilde{K}$ solutions are found.
\cref{alg:top-k-rel-neur-walk} summarizes this procedure.

\subsection{Upper Bound of Number of Subsets}
\label{app:subset-no}

Formally,
the search space splitting procedure, described in 
Appendix~\ref{sec:App.SearchSpaceSplitting},
generates $(\widetilde{K} L + 1)$
subsets in the worst case.
None of those subsets is empty
if $\widetilde K \leq M\underline{N}$, where $\underline{N} = \min N^{(l)}$,
while some can be empty otherwise.
Therefore, the number of subsets is upper-bounded by $\widetilde{K} L + 1$.

\subsection{Proof of \cref{thrm:ExactNeuronLevelSearch}: Complexity of Top-$\widetilde K$ Neuron-level Walk Search}
\label{app:topk-complexity}

Appendix~\ref{sec:DerivationEmpNeuBasic} and 
Appendix~\ref{sec:App.SearchSpaceSplitting} already explained how EMP-neu finds the top-$\widetilde K$ neuron-level walks with the highest absolute relevances,
and the top-$K$ walks with the highest positive relevances can be found from those solutions by assumption.
Below we consider the computational complexity.

According to Appendix~\ref{app:top1-complexity}, finding the walk with the highest absolute relevance by EMP-neu-Basic requires $\mcO(LM^2 \overline{N}^2)$ time.
For finding the second to the $\widetilde{K}$-th solution, the search space splitting generates no more than $(\widetilde{K} L + 1)$ subsets (see Appendix~\ref{app:subset-no}), for each of which EMP-neu-Basic needs to be applied.
However, in finding the $k$-th solution, maximization in each subset is operated effectively only at one layer with the messages $\widehat{m}_l(m_{l-1}, n_{l-1})$ and $\widehat{n}_l (m_{l-1}, n_{l-1})$ already computed when the top-$(k-1)$ solutions were searched.  Therefore, by reusing those messages, the computation complexity to find the maximizer from each subset for the $k (\geq 2)$-th solution search is only $\mcO(LM \overline{N}) $.
Therefore, the total computational cost of EMP-neu is  $\mcO(LM^2 \overline{N}^2  + \widetilde{K} L (LM \overline{N})) $.
The memory cost is dominated by the storage for the transition matrices (see Appendix~\ref{app:top1-complexity}) and thus $\mcO(LM^2 \overline{N}^2)$.
\qed

\begin{algorithm}[t]
   \caption{Find the most relevant node-level walk approximately by averaging (AMP-ave-Basic)}
   \label{alg:approx-most-rel-node-walk}
\begin{algorithmic}
   \STATE {\bfseries Input:} \# of nodes: $M$, LRP transition matrices $\{\boldsymbol T^{l, m_l, m_{l+1}}\}$,    initial messages $\{\widehat{\bfmu}^{L, m_L}\}$ such that $\widehat{\mu}_{n_{L}}^{L, m_L} = |r_{n_{L}}^{L, m_L}|$
   
   \FOR{$l=L$ {\bfseries to} $1$}
       \FOR{${m_{l-1}}=1$ {\bfseries to} $M$}
           \STATE Find ${m_l}$ by solving: 
           
           $\argmax_{{m_l}} \sum_{n_{l-1}}\sum_{n_l} T^{l-1, {m_{l-1}}, {m_{l}}}_{{n_{l-1}}, {n_{l}}} \widehat{\mu}^{l, {m_{l}}}_{{n_l}}$
           \STATE Store the result in a maximum step mapping ${m_{l-1}} \rightarrow m_{l}$.
           \STATE Compute the relevance along the corresponding maximum walk step for ${m_{l-1}}$:

           $\widehat{\mu}^{l-1, {m_{l-1}}}_{{n_{l-1}}} = \sum_{n_l} T^{l-1, {m_{l-1}}, {m_{l}}}_{{n_{l-1}}, {n_{l}}} \widehat{\mu}^{l, {m_{l}}}_{{n_l}}$
       \ENDFOR
   \ENDFOR

   \STATE Select $m_0^*$ by $\argmax_{m_0} \sum_{n_0} \widehat{\mu}^{0, {m_{0}}}_{{n_0}}$.
   \STATE Read from the maximum step mappings the full walk $\bfm^*=m_0^* \rightarrow m_1^* \rightarrow \cdots \rightarrow m_L^*$.
   \STATE {\bfseries return} $\bfm^*$.
\end{algorithmic}
\end{algorithm}

\begin{algorithm}[t]
   \caption{Search space splitting for approximately finding the top-$\widetilde{K}$ most relevant node-level walks}
   \label{alg:approx-top-k-rel-node-walk}
\begin{algorithmic}
   \STATE {\bfseries Input:} top-1 most relevant node-level walk $\bfm^1$, maximum step mappings $m_{l-1} \rightarrow m_{l}$.
   \STATE Initialize $SearchSpace = \{\}$
   \STATE Initialize $TopKWalks = \{\bfm^1\}$

   \FOR{$l=0$ {\bfseries to} $L$}
       \STATE $subset = \{\bfm: m_j= m_j^{1} \ \forall{j<l}, m_l \ne m_l^{1})\}$.
            \STATE Set the beginning of $\bfm^*$ such that $m_j^* = m_j^{1} \ \forall j<l$.
            \STATE Read from the maximum step mappings the following steps $m^*_l \rightarrow \cdots \rightarrow m^*_L$.
            \STATE Add $(subset: \bfm^*)$ to $SearchSpace$.
    
    \ENDFOR
   \FOR{$k=2$ {\bfseries to} $\widetilde{K}$}

        \STATE $MaxRelevance = -\infty$
        \FOR{\{$subset: \bfm^*\}$ {\bfseries in} $SearchSpace$}
            \IF{ $R^{\bfm^*} > MaxRelevance$} 
                \STATE $MaxRelevance = R^{\bfm^*}$.
                \STATE $\bfm^k = \bfm^*$.
                \STATE $MaxSubset = subset$.
            \ENDIF
        \ENDFOR

        \STATE Add $\bfm^k$ to $TopKWalks$.
        \STATE Split $MaxSubset$ similar to \eqref{eq:top-3-neuron-walk.app}, read out each subspace' most relevant walk, and add them to $SearchSpace$.
        \STATE Remove $MaxSubset$ from $SearchSpace$.
   \ENDFOR
   \STATE {\bfseries return} $TopKWalks$.
\end{algorithmic}
\end{algorithm}

\section{Details of Approximate Max-product Message Passing Algorithm with Averaging for Node-level Walks (AMP-ave)}
\label{sec:DerivationNodeLevelApproximateMaxProduct}

\cref{alg:approx-most-rel-node-walk} describes the detailed steps of {\bf AMP-ave-Basic} that approximately finds the most relevant node-level walk,
while \cref{alg:approx-top-k-rel-node-walk} describes the steps of search space splitting for the top-$\widetilde{K}$ node level search.

Similarly to EMP-neu-Basic, AMP-neu-Basic (\cref{alg:approx-most-rel-node-walk}) requires  $\mathcal O(LM^2N^2)$ time,
and the search space splitting (\cref{alg:approx-top-k-rel-node-walk}) requires 
$\mathcal O(KL^2MN)$, and thus the total computational complexity is $\mathcal O(LM^2N^2 + KL^2MN)$.

A naive implementation requires the same memory cost $\mathcal O(LM^2N^2)$ as EMP-neu to store the transition matrices.
However, there is a way to reduce the memory cost to $\mathcal O(M^2+LN(M+N)) = \mcO(L \max(M, \overline{N})^2)$, as explained below.

\subsection{Low-Memory Computation for Top-$K$ Node-level Walk Algorithm}
\label{sec:deri-no-T}

Substituting the explicit expression 
\eqref{eq:TforLRPGamma} of the propagation matrices
for GCN into the objective in the message passing \eqref{eq:MaxProductNodeLevelWalk}
gives
\begin{equation}
    \begin{aligned}
   \textstyle     \sum_{n_{l-1}}\sum_{n_l} T^{l-1, m_{l-1}, m_{l}}_{n_{l-1}, n_{l}} \widehat{\mu}^{l, m_{l}}_{n_l} 
     &= \textstyle  \sum_{n_{l-1}}\sum_{n_l} \frac{\Lambda_{m_{l-1}, m_{l}} H_{m_{l-1}, n_{l-1}}^{(l-1)} W^{\uparrow(l-1)}_{n_{l-1}, n_{l}} }{\sum_{m'' ,n''} {\Lambda_{m'', m_{l}} H_{m'', n''}^{(l-1)} W^{\uparrow(l-1)}_{n'', n_{l} } }} \widehat{\mu}^{l, m_{l}}_{n_l} \\
     &= \textstyle  \Lambda_{m_{l-1}, m_{l}} \sum_{n_l} \frac{ \boldsymbol H_{m_{l-1}}^{(l-1)}  \boldsymbol W^{\uparrow(l-1)}_{n_{l}}}{\sum_{m''} {\Lambda_{m'', m_{l}} \boldsymbol H_{m''}^{(l-1)}  \boldsymbol W^{\uparrow(l-1)}_{n_{l}} }} \widehat{\mu}^{l, m_{l}}_{n_l},
     \label{eq:LowCostExpression}
    \end{aligned}
\end{equation}
where $\bfLambda \in \mathbb{R}^{M \times M}, \bfH^{(l)} \in \mathbb R^{M\times N^{(l)}}, \bfW^{\uparrow (l)} \in \mathbb R^{N^{(l-1)} \times N^{(l)}}$
are the adjacency matrix, 
the activation matrix, and 
the modified weight matrix, respectively
(see Sections~\ref{sec:GraphNeuralNetworks} and \ref{sec:WalkRelevance}).
The expression \eqref{eq:LowCostExpression} implies that the maximization \eqref{eq:MaxProductNodeLevelWalk}, as well as the relevance message passing \eqref{eq:MaxProductNodeLevelMessage},
can be performed by using 
$\bfLambda,\{ \boldsymbol{H}^{(l)}, \boldsymbol{W}^{\uparrow (l)}\}$,
which requires $\mcO(M^2 + L M \overline{N}+ L\overline{N}^2)$ memory, without storing the  transition matrices $\{\boldsymbol T^{(l), m_{l}, m_{l+1}}\}$, which requires $\mathcal O(M^2 \overline{N}^2)$ memory.
The memory cost of this implementation is thus 
$\mcO(L \max(M, \overline{N})^2)$.

\section{Empirical Investigation of Algorithm Behavior}

\subsection{Proportion of Positive Relevant Walks in Top-$\widetilde K$ Absolute Relevant Walks}
\label{sec:NecessaryKTilde}

\begin{figure}
\begin{center}
\centerline{\includegraphics[width=0.35\linewidth]{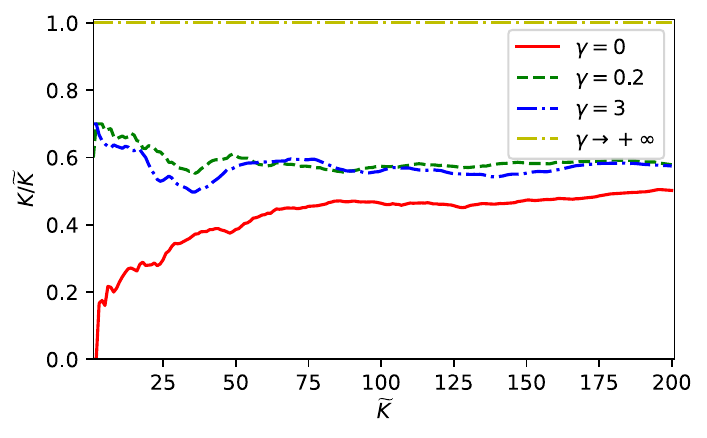}}
\caption{The proportion $\frac{K}{\widetilde K}$ of positive relevant neuron-level walks in the top-$\widetilde K$ absolute relevant walks.}
\label{fig:widetilde-K-K}
\end{center}
\vskip -0.3in
\end{figure}

We randomly chose 10 correctly classified graph samples from MUTAG and found the top-$\widetilde K$ absolute relevant neuron-level walks by EMP-neu.
\cref{fig:widetilde-K-K} plots the proportion
$\frac{K}{\widetilde K}$ of the number $K$ of positive relevant walks in the top-$\widetilde K$ absolute relevant walks.
We observe that higher $\gamma$ leads to a larger proportion of positive walks,
and that more than half walks are positive for $\gamma \geq 0.2$.

\subsection{Sign of Maximization Objective in Eq.\eqref{eq:MaxProductNodeLevelWalk}}
\label{app:max-obj-positiv}

\cref{tab:max-obj-positiv} shows how often the objective of the maximization objective in the AMP-ave message passing \eqref{eq:MaxProductNodeLevelWalk}
is positive, negative, or zero on randomly chosen 10 correctly classified samples from BA-2motif, MUTAG, and Graph-SST2 with different $\gamma$.
We see that the objective tends to be positive or zero for $\gamma \geq 0.2$, although it can be negative with significant frequency, compared to the positive frequency, in BA-2motif.

\begin{table}
    \centering
    \begin{tabular}{clrrr}
        \toprule
       Dataset  & $\gamma$ & $+$ & $-$ & 0\\
       \midrule
        BA-2motif & 0 & 3.70\% & 4.12\% & 92.10\% \\
         & 0.2 & 6.38\% & 2.78\% & 90.84\% \\
         & $[3,\dots,0]$ & 5.34\% & 2.66\% & 92.01\% \\
         & $+\infty$ & 6.70\% & 2.67\% & 90.63\% \\
        MUTAG & 0 & 11.82\% & 2.48\% & 85.70\% \\
         & 0.2 & 12.39\% & 2.23\% & 85.38\% \\
         & $[3,\dots,0]$ & 12.86\% & 1.95\% & 85.19\% \\
         & $+\infty$ & 14.28\% & 1.12\% & 84.60\% \\
        Graph-SST2 & 0 & 39.17\% & 3.30\% & 57.53\% \\
         & 0.2 & 41.45\% & 2.41\% & 56.14\% \\
         & $[3,\dots,0]$ & 42.06\% & 1.76\% & 56.18\% \\
         & $+\infty$ & 40.48\% & 3.30\% & 56.22\% \\
         \bottomrule
    \end{tabular}
    \caption{Frequency of the sign of the objective in AMP-ave message passing 
 \eqref{eq:MaxProductNodeLevelWalk}.}
    \label{tab:max-obj-positiv}
\end{table}

\subsection{Empirical results from EMP-neu}
\label{app:emp-neu-verify}

Here
we verify the correctness of the top neuron-level walks found by EMP-neu.
For one graph each from the positive (\cref{tab:emp-neu-walks-pos}) and the negative (\cref{tab:emp-neu-walks}) classes in BA-2motif, we list the top 100 most absolute relevant walks (before omitting the negative-relevant walks) found by EMP-neu. 
The first and the second columns, respectively, show the estimated ranking by EMP-neu and the true ranking found by the exhaustive search.
We observe that the true ranking is in non-decreasing order, proving the correctness of EMP-neu.
Notably, our method took only 3.4 seconds to find the top 100 most absolute relevant walks, while the exhaustive search took over 3 hours, because the number of possible neuron-level walks is $\mathcal O((M\overline N)^L)$.

\begin{table}
    \centering
    \small
    \begin{tabular}{rrrr|rrrr}
        \toprule
       \makecell{Est. \\ \#} & \makecell{True \\ \#} & Most absolute relevant walks & Relevance & \makecell{Est. \\ \#} & \makecell{True \\ \#} & Most absolute relevant walks &  Relevance\\
       \midrule
        1       & 1       & ((21, 0), (22, 9), (23, 11), (24, 0)) & 0.12953492 & 51      & 10      & ((23, 0), (23, 9), (22, 11), (22, 0)) & 0.12953489 \\
2       & 1       & ((22, 0), (22, 9), (23, 11), (24, 0)) & 0.12953492 & 52      & 10      & ((23, 0), (23, 9), (22, 11), (23, 0)) & 0.12953489 \\
3       & 1       & ((23, 0), (22, 9), (23, 11), (24, 0)) & 0.12953492 & 53      & 10      & ((20, 0), (21, 9), (22, 11), (22, 0)) & 0.12953489 \\
4       & 1       & ((24, 0), (23, 9), (23, 11), (24, 0)) & 0.12953492 & 54      & 10      & ((20, 0), (21, 9), (22, 11), (23, 0)) & 0.12953489 \\
5       & 1       & ((22, 0), (23, 9), (23, 11), (24, 0)) & 0.12953492 & 55      & 55      & ((0, 0), (13, 9), (15, 11), (13, 0))  & 0.10064131 \\
6       & 1       & ((23, 0), (23, 9), (23, 11), (24, 0)) & 0.12953492 & 56      & 55      & ((13, 0), (13, 9), (15, 11), (13, 0)) & 0.10064131 \\
7       & 7       & ((20, 0), (24, 9), (23, 11), (24, 0)) & 0.12953490 & 57      & 55      & ((15, 0), (13, 9), (15, 11), (13, 0)) & 0.10064131 \\
8       & 7       & ((23, 0), (24, 9), (23, 11), (24, 0)) & 0.12953490 & 58      & 55      & ((19, 0), (15, 9), (15, 11), (13, 0)) & 0.10064131 \\
9       & 7       & ((24, 0), (24, 9), (23, 11), (24, 0)) & 0.12953490 & 59      & 55      & ((0, 0), (13, 9), (15, 11), (15, 0))  & 0.10064131 \\
10      & 10      & ((21, 0), (21, 9), (22, 11), (21, 0)) & 0.12953489 & 60      & 55      & ((0, 0), (13, 9), (15, 11), (19, 0))  & 0.10064131 \\
11      & 10      & ((21, 0), (22, 9), (22, 11), (21, 0)) & 0.12953489 & 61      & 55      & ((13, 0), (15, 9), (15, 11), (13, 0)) & 0.10064131 \\
12      & 10      & ((21, 0), (22, 9), (23, 11), (22, 0)) & 0.12953489 & 62      & 55      & ((13, 0), (13, 9), (15, 11), (15, 0)) & 0.10064131 \\
13      & 10      & ((21, 0), (22, 9), (23, 11), (23, 0)) & 0.12953489 & 63      & 55      & ((13, 0), (13, 9), (15, 11), (19, 0)) & 0.10064131 \\
14      & 10      & ((22, 0), (21, 9), (22, 11), (21, 0)) & 0.12953489 & 64      & 55      & ((15, 0), (15, 9), (15, 11), (13, 0)) & 0.10064131 \\
15      & 10      & ((22, 0), (22, 9), (22, 11), (21, 0)) & 0.12953489 & 65      & 55      & ((15, 0), (13, 9), (15, 11), (15, 0)) & 0.10064131 \\
16      & 10      & ((22, 0), (22, 9), (23, 11), (22, 0)) & 0.12953489 & 66      & 55      & ((15, 0), (13, 9), (15, 11), (19, 0)) & 0.10064131 \\
17      & 10      & ((22, 0), (22, 9), (23, 11), (23, 0)) & 0.12953489 & 67      & 55      & ((19, 0), (15, 9), (15, 11), (15, 0)) & 0.10064131 \\
18      & 10      & ((23, 0), (22, 9), (22, 11), (21, 0)) & 0.12953489 & 68      & 55      & ((19, 0), (15, 9), (15, 11), (19, 0)) & 0.10064131 \\
19      & 10      & ((23, 0), (22, 9), (23, 11), (22, 0)) & 0.12953489 & 69      & 55      & ((13, 0), (15, 9), (15, 11), (15, 0)) & 0.10064131 \\
20      & 10      & ((23, 0), (22, 9), (23, 11), (23, 0)) & 0.12953489 & 70      & 55      & ((13, 0), (15, 9), (15, 11), (19, 0)) & 0.10064131 \\
21      & 10      & ((24, 0), (23, 9), (22, 11), (21, 0)) & 0.12953489 & 71      & 55      & ((15, 0), (15, 9), (15, 11), (15, 0)) & 0.10064131 \\
22      & 10      & ((24, 0), (23, 9), (23, 11), (22, 0)) & 0.12953489 & 72      & 55      & ((15, 0), (15, 9), (15, 11), (19, 0)) & 0.10064131 \\
23      & 10      & ((24, 0), (23, 9), (23, 11), (23, 0)) & 0.12953489 & 73      & 73      & ((21, 0), (20, 9), (24, 11), (24, 0)) & 0.07078890 \\
24      & 10      & ((22, 0), (23, 9), (22, 11), (21, 0)) & 0.12953489 & 74      & 73      & ((24, 0), (20, 9), (24, 11), (24, 0)) & 0.07078890 \\
25      & 10      & ((22, 0), (23, 9), (23, 11), (22, 0)) & 0.12953489 & 75      & 73      & ((20, 0), (20, 9), (24, 11), (24, 0)) & 0.07078890 \\
26      & 10      & ((22, 0), (23, 9), (23, 11), (23, 0)) & 0.12953489 & 76      & 73      & ((0, 0), (20, 9), (24, 11), (24, 0))  & 0.07078890 \\
27      & 10      & ((23, 0), (23, 9), (22, 11), (21, 0)) & 0.12953489 & 77      & 77      & ((21, 0), (20, 9), (21, 11), (21, 0)) & 0.07078888 \\
28      & 10      & ((23, 0), (23, 9), (23, 11), (22, 0)) & 0.12953489 & 78      & 77      & ((21, 0), (20, 9), (24, 11), (23, 0)) & 0.07078888 \\
29      & 10      & ((23, 0), (23, 9), (23, 11), (23, 0)) & 0.12953489 & 79      & 77      & ((24, 0), (20, 9), (21, 11), (21, 0)) & 0.07078888 \\
30      & 10      & ((20, 0), (21, 9), (22, 11), (21, 0)) & 0.12953489 & 80      & 77      & ((24, 0), (20, 9), (24, 11), (23, 0)) & 0.07078888 \\
31      & 10      & ((20, 0), (24, 9), (23, 11), (22, 0)) & 0.12953489 & 81      & 77      & ((20, 0), (20, 9), (21, 11), (21, 0)) & 0.07078888 \\
32      & 10      & ((20, 0), (24, 9), (23, 11), (23, 0)) & 0.12953489 & 82      & 77      & ((20, 0), (20, 9), (24, 11), (23, 0)) & 0.07078888 \\
33      & 10      & ((23, 0), (24, 9), (23, 11), (22, 0)) & 0.12953489 & 83      & 77      & ((0, 0), (20, 9), (21, 11), (21, 0))  & 0.07078888 \\
34      & 10      & ((23, 0), (24, 9), (23, 11), (23, 0)) & 0.12953489 & 84      & 77      & ((0, 0), (20, 9), (24, 11), (23, 0))  & 0.07078888 \\
35      & 10      & ((24, 0), (24, 9), (23, 11), (22, 0)) & 0.12953489 & 85      & 85      & ((21, 0), (20, 9), (21, 11), (22, 0)) & 0.07078888 \\
36      & 10      & ((24, 0), (24, 9), (23, 11), (23, 0)) & 0.12953489 & 86      & 85      & ((24, 0), (20, 9), (21, 11), (22, 0)) & 0.07078888 \\
37      & 10      & ((21, 0), (21, 9), (22, 11), (22, 0)) & 0.12953489 & 87      & 85      & ((20, 0), (20, 9), (21, 11), (22, 0)) & 0.07078888 \\
38      & 10      & ((21, 0), (21, 9), (22, 11), (23, 0)) & 0.12953489 & 88      & 85      & ((0, 0), (20, 9), (21, 11), (22, 0))  & 0.07078888 \\
39      & 10      & ((21, 0), (22, 9), (22, 11), (22, 0)) & 0.12953489 & 89      & 89      & ((21, 0), (20, 9), (24, 11), (20, 0)) & 0.07078885 \\
40      & 10      & ((21, 0), (22, 9), (22, 11), (23, 0)) & 0.12953489 & 90      & 89      & ((24, 0), (20, 9), (24, 11), (20, 0)) & 0.07078885 \\
41      & 10      & ((22, 0), (21, 9), (22, 11), (22, 0)) & 0.12953489 & 91      & 89      & ((20, 0), (20, 9), (24, 11), (20, 0)) & 0.07078885 \\
42      & 10      & ((22, 0), (21, 9), (22, 11), (23, 0)) & 0.12953489 & 92      & 89      & ((0, 0), (20, 9), (24, 11), (20, 0))  & 0.07078885 \\
43      & 10      & ((22, 0), (22, 9), (22, 11), (22, 0)) & 0.12953489 & 93      & 93      & ((21, 0), (20, 9), (21, 11), (20, 0)) & 0.07078885 \\
44      & 10      & ((22, 0), (22, 9), (22, 11), (23, 0)) & 0.12953489 & 94      & 93      & ((24, 0), (20, 9), (21, 11), (20, 0)) & 0.07078885 \\
45      & 10      & ((23, 0), (22, 9), (22, 11), (22, 0)) & 0.12953489 & 95      & 93      & ((20, 0), (20, 9), (21, 11), (20, 0)) & 0.07078885 \\
46      & 10      & ((23, 0), (22, 9), (22, 11), (23, 0)) & 0.12953489 & 96      & 93      & ((0, 0), (20, 9), (21, 11), (20, 0))  & 0.07078885 \\
47      & 10      & ((24, 0), (23, 9), (22, 11), (22, 0)) & 0.12953489 & 97      & 97      & ((1, 0), (2, 9), (10, 11), (2, 0))    & 0.06455576 \\
48      & 10      & ((24, 0), (23, 9), (22, 11), (23, 0)) & 0.12953489 & 98      & 97      & ((2, 0), (2, 9), (10, 11), (2, 0))    & 0.06455576 \\
49      & 10      & ((22, 0), (23, 9), (22, 11), (22, 0)) & 0.12953489 & 99      & 97      & ((10, 0), (2, 9), (10, 11), (2, 0))   & 0.06455576 \\
50      & 10      & ((22, 0), (23, 9), (22, 11), (23, 0)) & 0.12953489 & 100     & 97      & ((1, 0), (2, 9), (10, 11), (10, 0))   & 0.06455576 \\ 
        
        \bottomrule
    \end{tabular}
    \caption{The top 100 most absolute relevant node-level walks (found by EMP-neu).  The columns show the estimated ranking by EMP-neu, the true ranking found by the exhaustive search, the neuron-level walk, and its relevance. The walk is expressed as $L + 1 = 4$ steps of node-neuron pairs $(m,n)$, denoting the $n$-th neuron of the $m$-th node. The graph is from the positive class of BA-2motif dataset. Note that the walks with the same relevance share the same ranking.}
    \label{tab:emp-neu-walks-pos}
\end{table}

\begin{table}
    \centering
    \small
    \begin{tabular}{rrrr|rrrr}
        \toprule
       \makecell{Est. \\ \#} & \makecell{True \\ \#} & Most absolute relevant walks & Relevance & \makecell{Est. \\ \#} & \makecell{True \\ \#} & Most absolute relevant walks &  Relevance\\
       \midrule
        1  & 1  & ((20, 0), (21, 9), (22, 11), (23, 1)) & -0.062610321 & 51  & 49 & ((13, 0), (7, 9), (7, 11), (7, 1))   & -0.039949972 \\
2  & 1  & ((21, 0), (21, 9), (22, 11), (23, 1)) & -0.062610321 & 52  & 49 & ((15, 0), (7, 9), (7, 11), (7, 1))   & -0.039949972 \\
3  & 1  & ((22, 0), (21, 9), (22, 11), (23, 1)) & -0.062610321 & 53  & 53 & ((5, 0), (7, 9), (7, 11), (5, 1))    & -0.039949801 \\
4  & 1  & ((24, 0), (21, 9), (22, 11), (23, 1)) & -0.062610321 & 54  & 53 & ((5, 0), (7, 9), (7, 11), (13, 1))   & -0.039949801 \\
5  & 5  & ((20, 0), (21, 9), (22, 11), (21, 1)) & -0.062610313 & 55  & 53 & ((5, 0), (7, 9), (7, 11), (15, 1))   & -0.039949801 \\
6  & 5  & ((20, 0), (21, 9), (22, 11), (22, 1)) & -0.062610313 & 56  & 53 & ((7, 0), (7, 9), (7, 11), (5, 1))    & -0.039949801 \\
7  & 5  & ((21, 0), (21, 9), (22, 11), (21, 1)) & -0.062610313 & 57  & 53 & ((7, 0), (7, 9), (7, 11), (13, 1))   & -0.039949801 \\
8  & 5  & ((21, 0), (21, 9), (22, 11), (22, 1)) & -0.062610313 & 58  & 53 & ((7, 0), (7, 9), (7, 11), (15, 1))   & -0.039949801 \\
9  & 5  & ((22, 0), (21, 9), (22, 11), (21, 1)) & -0.062610313 & 59  & 53 & ((13, 0), (7, 9), (7, 11), (5, 1))   & -0.039949801 \\
10 & 5  & ((22, 0), (21, 9), (22, 11), (22, 1)) & -0.062610313 & 60  & 53 & ((13, 0), (7, 9), (7, 11), (13, 1))  & -0.039949801 \\
11 & 5  & ((24, 0), (21, 9), (22, 11), (21, 1)) & -0.062610313 & 61  & 53 & ((13, 0), (7, 9), (7, 11), (15, 1))  & -0.039949801 \\
12 & 5  & ((24, 0), (21, 9), (22, 11), (22, 1)) & -0.062610313 & 62  & 53 & ((15, 0), (7, 9), (7, 11), (5, 1))   & -0.039949801 \\
13 & 13 & ((2, 0), (9, 9), (16, 11), (16, 1))   & -0.057097323 & 63  & 53 & ((15, 0), (7, 9), (7, 11), (13, 1))  & -0.039949801 \\
14 & 13 & ((4, 0), (8, 9), (12, 11), (8, 1))    & -0.057097323 & 64  & 53 & ((15, 0), (7, 9), (7, 11), (15, 1))  & -0.039949801 \\
15 & 13 & ((8, 0), (8, 9), (12, 11), (8, 1))    & -0.057097323 & 65  & 65 & ((0, 0), (1, 9), (2, 5), (2, 1))     & 0.037394594  \\
16 & 13 & ((9, 0), (9, 9), (16, 11), (16, 1))   & -0.057097323 & 66  & 65 & ((1, 0), (1, 9), (2, 5), (2, 1))     & 0.037394594  \\
17 & 13 & ((12, 0), (8, 9), (12, 11), (8, 1))   & -0.057097323 & 67  & 65 & ((3, 0), (1, 9), (2, 5), (2, 1))     & 0.037394594  \\
18 & 13 & ((14, 0), (14, 9), (19, 11), (14, 1)) & -0.057097323 & 68  & 65 & ((18, 0), (1, 9), (2, 5), (2, 1))    & 0.037394594  \\
19 & 13 & ((16, 0), (9, 9), (16, 11), (16, 1))  & -0.057097323 & 69  & 65 & ((20, 0), (20, 9), (21, 5), (20, 1)) & 0.037394594  \\
20 & 13 & ((19, 0), (14, 9), (19, 11), (14, 1)) & -0.057097323 & 70  & 65 & ((21, 0), (20, 9), (21, 5), (20, 1)) & 0.037394594  \\
21 & 13 & ((4, 0), (14, 9), (19, 11), (14, 1))  & -0.057097323 & 71  & 65 & ((24, 0), (20, 9), (21, 5), (20, 1)) & 0.037394594  \\
22 & 13 & ((4, 0), (8, 9), (12, 11), (12, 1))   & -0.057097323 & 72  & 65 & ((2, 0), (1, 9), (2, 5), (2, 1))     & 0.037394594  \\
23 & 13 & ((8, 0), (8, 9), (12, 11), (12, 1))   & -0.057097323 & 73  & 65 & ((23, 0), (20, 9), (21, 5), (20, 1)) & 0.037394594  \\
24 & 13 & ((12, 0), (8, 9), (12, 11), (12, 1))  & -0.057097323 & 74  & 65 & ((0, 0), (20, 9), (21, 5), (20, 1))  & 0.037394594  \\
25 & 13 & ((14, 0), (14, 9), (19, 11), (19, 1)) & -0.057097323 & 75  & 65 & ((0, 0), (1, 9), (2, 5), (9, 1))     & 0.037394594  \\
26 & 13 & ((19, 0), (14, 9), (19, 11), (19, 1)) & -0.057097323 & 76  & 65 & ((1, 0), (1, 9), (2, 5), (9, 1))     & 0.037394594  \\
27 & 13 & ((4, 0), (14, 9), (19, 11), (19, 1))  & -0.057097323 & 77  & 65 & ((3, 0), (1, 9), (2, 5), (9, 1))     & 0.037394594  \\
28 & 28 & ((2, 0), (9, 9), (16, 11), (9, 1))    & -0.057097316 & 78  & 65 & ((18, 0), (1, 9), (2, 5), (9, 1))    & 0.037394594  \\
29 & 28 & ((9, 0), (9, 9), (16, 11), (9, 1))    & -0.057097316 & 79  & 65 & ((20, 0), (20, 9), (21, 5), (24, 1)) & 0.037394594  \\
30 & 28 & ((16, 0), (9, 9), (16, 11), (9, 1))   & -0.057097316 & 80  & 65 & ((21, 0), (20, 9), (21, 5), (24, 1)) & 0.037394594  \\
31 & 31 & ((23, 0), (22, 9), (22, 11), (23, 1)) & -0.045452297 & 81  & 65 & ((24, 0), (20, 9), (21, 5), (24, 1)) & 0.037394594  \\
32 & 31 & ((20, 0), (23, 9), (22, 11), (23, 1)) & -0.045452297 & 82  & 65 & ((2, 0), (1, 9), (2, 5), (9, 1))     & 0.037394594  \\
33 & 31 & ((21, 0), (22, 9), (22, 11), (23, 1)) & -0.045452297 & 83  & 65 & ((23, 0), (20, 9), (21, 5), (24, 1)) & 0.037394594  \\
34 & 31 & ((22, 0), (22, 9), (22, 11), (23, 1)) & -0.045452297 & 84  & 65 & ((0, 0), (20, 9), (21, 5), (24, 1))  & 0.037394594  \\
35 & 31 & ((22, 0), (23, 9), (22, 11), (23, 1)) & -0.045452297 & 85  & 85 & ((0, 0), (1, 9), (2, 5), (1, 1))     & 0.037394591  \\
36 & 31 & ((23, 0), (23, 9), (22, 11), (23, 1)) & -0.045452297 & 86  & 85 & ((0, 0), (1, 9), (2, 5), (5, 1))     & 0.037394591  \\
37 & 37 & ((23, 0), (22, 9), (22, 11), (21, 1)) & -0.045452293 & 87  & 85 & ((1, 0), (1, 9), (2, 5), (1, 1))     & 0.037394591  \\
38 & 37 & ((23, 0), (22, 9), (22, 11), (22, 1)) & -0.045452293 & 88  & 85 & ((1, 0), (1, 9), (2, 5), (5, 1))     & 0.037394591  \\
39 & 37 & ((20, 0), (23, 9), (22, 11), (21, 1)) & -0.045452293 & 89  & 85 & ((3, 0), (1, 9), (2, 5), (1, 1))     & 0.037394591  \\
40 & 37 & ((20, 0), (23, 9), (22, 11), (22, 1)) & -0.045452293 & 90  & 85 & ((3, 0), (1, 9), (2, 5), (5, 1))     & 0.037394591  \\
41 & 37 & ((21, 0), (22, 9), (22, 11), (21, 1)) & -0.045452293 & 91  & 85 & ((18, 0), (1, 9), (2, 5), (1, 1))    & 0.037394591  \\
42 & 37 & ((21, 0), (22, 9), (22, 11), (22, 1)) & -0.045452293 & 92  & 85 & ((18, 0), (1, 9), (2, 5), (5, 1))    & 0.037394591  \\
43 & 37 & ((22, 0), (22, 9), (22, 11), (21, 1)) & -0.045452293 & 93  & 85 & ((20, 0), (20, 9), (21, 5), (21, 1)) & 0.037394591  \\
44 & 37 & ((22, 0), (22, 9), (22, 11), (22, 1)) & -0.045452293 & 94  & 85 & ((20, 0), (20, 9), (21, 5), (22, 1)) & 0.037394591  \\
45 & 37 & ((22, 0), (23, 9), (22, 11), (21, 1)) & -0.045452293 & 95  & 85 & ((21, 0), (20, 9), (21, 5), (21, 1)) & 0.037394591  \\
46 & 37 & ((22, 0), (23, 9), (22, 11), (22, 1)) & -0.045452293 & 96  & 85 & ((21, 0), (20, 9), (21, 5), (22, 1)) & 0.037394591  \\
47 & 37 & ((23, 0), (23, 9), (22, 11), (21, 1)) & -0.045452293 & 97  & 85 & ((24, 0), (20, 9), (21, 5), (21, 1)) & 0.037394591  \\
48 & 37 & ((23, 0), (23, 9), (22, 11), (22, 1)) & -0.045452293 & 98  & 85 & ((24, 0), (20, 9), (21, 5), (22, 1)) & 0.037394591  \\
49 & 49 & ((5, 0), (7, 9), (7, 11), (7, 1))     & -0.039949972 & 99  & 85 & ((2, 0), (1, 9), (2, 5), (1, 1))     & 0.037394591  \\
50 & 49 & ((7, 0), (7, 9), (7, 11), (7, 1))     & -0.039949972 & 100 & 85 & ((2, 0), (1, 9), (2, 5), (5, 1))     & 0.037394591  \\
        
        \bottomrule
    \end{tabular}
    \caption{
    The top 100 most absolute relevant node-level walks (found by EMP-neu).  The columns show the estimated ranking by EMP-neu, the true ranking found by the exhaustive search, the neuron-level walk, and its relevance. The walk is expressed as $L + 1 = 4$ steps of node-neuron pairs $(m,n)$, denoting the $n$-th neuron of the $m$-th node. The graph is from the negative class of BA-2motif dataset. 
    Note that the walks with the same relevance share the same ranking.
    }
    \label{tab:emp-neu-walks}
\end{table}

\section{Additional Precision Evaluation on MUTAG and Graph-SST2}
\label{app:pr-graphsst2}

\begin{figure}[b]
\begin{center}
\centerline{\includegraphics[width=0.53\linewidth]{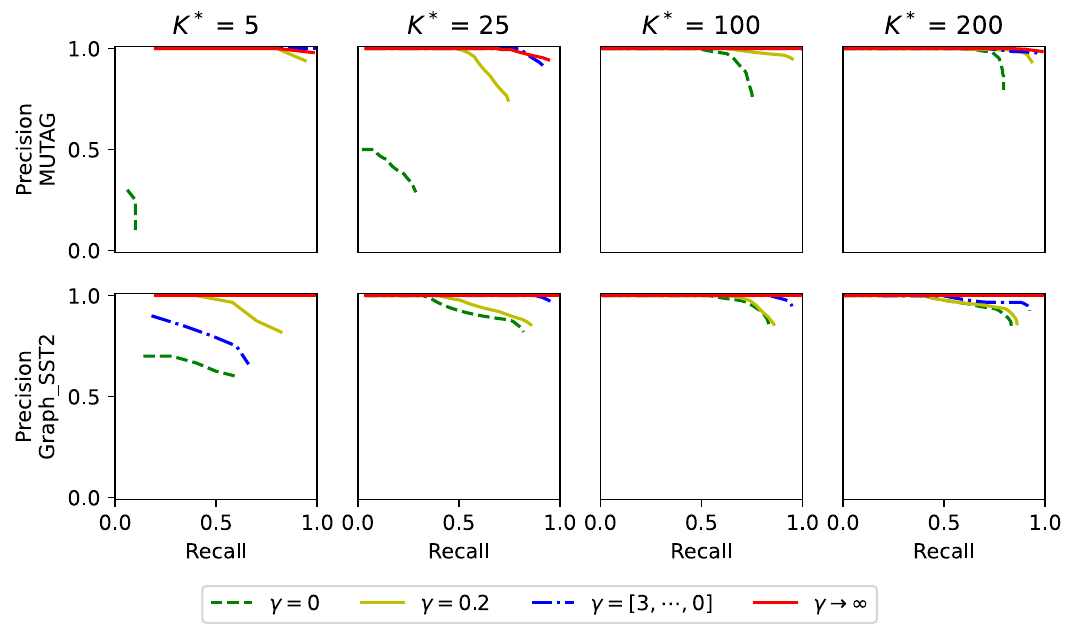}}
\vskip -0.1in
\caption{Precision-recall curves of AMP-ave for the top-$K^*$ node-level walks on MUTAG (top) and Graph-SST2 (bottom).
}
\label{fig:pr-top-k-sst2}
\end{center}
\vskip -0.1in
\end{figure}

\cref{fig:pr-top-k-sst2}
shows the precision-recall curves of AMP-ave on MUTAG and Graph-SST2.

\section{Datasets and GNN Models for Experiments}
\label{sec:models}

The datasets (summarized in \cref{tab:stat_datasets}) and models are mainly downloaded and trained according to the instructions provided in \citet{DBLP:conf/icml/XiongSMMN22}.

\subsection{BA-2motif}

BA-2motif \citep{luo2020parameterized} is a synthetic dataset of graphs that are built by attaching one of two different motifs (either a house or a circle shape) to a random graph, which is generated by the Barab\'asi-Albert (BA) model. The prediction task is to classify the graphs according to the motif type.

We trained a GIN model with 3 layers, with a 2-layer multi-layer perceptron as the combine function in every GIN block. The activation function we use throughout the model is ReLU. The nodes' initial embedding is single value 1 ($N^{(0)} = 1$), and the first 2 GIN blocks transforms the features into 20-dimensional and the last block transforms the feature into 2 dimensions. In the last layer we do a sum pooling across  all node features to get a vector that represents the whole graph, and transform it to class probabilities by a softmax function.
We trained the model with the SGD optimizer with a decreasing learning rate $\gamma = 0.00001 / (1.0 + (\texttt{epoch} / \texttt{epochs})$ for 5000 epochs. The dataset  consists of 1000 samples (500 for each class). We train the model with a set of 400 positive and 400 negative samples, and use the rest as test set. The test accuracy is 100\%.

\subsection{MUTAG}

MUTAG \citep{debnath1991structure} is a molecule dataset. The molecules are represented as graphs of atoms being nodes and chemical bonds being edges. 
The initial node features are one-hot vectors of atom types.
The samples are labelled as mutagenic and non-mutagenic.

For this dataset we apply a 3-layered GIN model, with 2-layer MLP as the combine function in each GIN block. The input node feature dimension is 7, and the output dimension of first 2 GIN blocks is 128, and the final block transforms the feature into 2 dimensions. The readout function is sum over all nodes followed by a softmax function.
The train set consists of 108 samples with half positive and half negative, and we use the rest samples as the test set. We trained the model with SGD optimizer for 1500 epochs, and the learning rate $\gamma = 0.0005 / (1.0 + (\texttt{epoch} / \texttt{epochs})$. The test accuracy is 85.00\%.

\begin{table*} [t]
    \caption{Statistics of the datasets.  }
    \label{tab:stat_datasets}
    \begin{center}
        \begin{small}
            \begin{sc}
                \begin{tabular}{lrrrrrrrrcc}
                    \toprule
                    & \multicolumn{2}{c}{BA-2motif} & \multicolumn{2}{c}{MUTAG} & \multicolumn{2}{c}{Mutagenicity} & \multicolumn{2}{c}{Graph-SST2} & %
                    Infection  \\
                    \midrule
                    \# of Edges (avg) & \multicolumn{2}{c}{25.48} & \multicolumn{2}{c}{19.79} & \multicolumn{2}{c}{17.79} & \multicolumn{2}{c}{19.40} & %
                    3991.30\\
                    \# of Nodes (avg) & \multicolumn{2}{c}{25.00} & \multicolumn{2}{c}{17.93} & \multicolumn{2}{c}{16.90} & \multicolumn{2}{c}{10.20} & %
                    1000   \\
                    \# of Graphs & \multicolumn{2}{c}{1000} & \multicolumn{2}{c}{188} & \multicolumn{2}{c}{4337} & \multicolumn{2}{c}{70042} & 
                    100     \\ 
                    \bottomrule
                \end{tabular}
            \end{sc}
        \end{small}
    \end{center}
    \vskip -0.1in
\end{table*} 

\subsection{Mutagenicity}\label{app:mutagenicity}

Mutagenicity \citep{doi:10.1021/jm040835a} is another bigger dataset for mutagenic and non-mutagenic molecules, and contains larger variety of types of mutagenic groups.

The model's input feature size is 13, and the rest settings are the same as used in MUTAG model.
The train set has 3096 samples with half positive and half negative, and the rest are used as test set. We trained the model with Adam optimizer for 25 epochs, and the initial learning rate $\gamma = 0.00005$. The test accuracy is 83.16\%.

\subsection{Graph-SST2}

Graph-SST2 \citep{yuan2020explainability} is a dataset of texts in the parse tree form, which is also represented as graphs. The node features are 768-dimensional word embedding vectors, which are pretrained and provided by the authors.

The model is built with a node feature embedding part and a following 3-layer GCN. The input feature dimension is 768, and in the middle layer of GCN the output dimension is 20. We downloaded the dataset from \citet{yuan2020explainability} and used their dataset split. We trained the model with Adam optimizer for 50 epochs, and the initial learning rate $\gamma = 0.0001$. The test accuracy is 89.40\%.

\section{Oracle Predictors for Infection Dataset}
\label{app:oracle}

We call \emph{oracle} who has the complete information about data generation process, and therefore provides predictions with best possible accuracy.
Since the analytic expressions of the predictive probabilities for the SI model are intractable, we estimate the probabilities by Monte Carlo sampling.

Assume that we fixed the parameters of the SI model, and drew a graph and initial carriers,
for which we will compute the oracle predictors. 
We simulate the infection process $Q=1000$ times and record the results.  We use a counter $x(m)$ to record how many times the node $m$ was infected after $L$ time steps.
We also use a counter $y(c^{(m)})$ to record how many times a particular infection chain $c^{(m)} \in \mcC^{(m)}$  occurred.
Here $\mcC^{(m)}$ denotes the set of possible infection chains from one of the initial carriers to the target node $m$.
With those records, we can estimate the following probabilities:
\begin{align}
\mathbb{P} (\text{Node $m$ will be infected})
& \approx \frac{x(m)}{Q},
\label{eq:OraclePredictionLabel}\\
\mathbb{P} (\text{Infection chain $c^{(m)}$ will occur})
&\approx \frac{y(c^{(m)})}{Q}.
\label{eq:OraclePredictionInfectionChain}
\end{align}

The estimators above converge to the true probabilities as $Q \to \infty$.
The oracle infection predictor based on Eq.\eqref{eq:OraclePredictionLabel} gives an accuracy upper-bound for any machine learning predictor,
while the oracle infection chain predictor based on  Eq.\eqref{eq:OraclePredictionInfectionChain} gives an upper-bound of the possible infection chain detection accuracy.  We can also obtain the set of possible infection chains by collecting chains such that Eq.\eqref{eq:OraclePredictionInfectionChain} is positive.

\section{Heuristics to Find the Most Relevant Walks based on Edge-IG}
\label{app:edge-ig-walk}

We rely on 
a natural assumption---the edges of which the most relevant walks consist should be relevant---and build the following heuristics.
We first sort the edges in descending order of the relevance scores given by 
Edge-IG. 
Then, we take the top-$\widetilde{K}$ edges
${(m,m')^{k}, k=1,\cdots,\widetilde{K}}$ with their relevance scores $R^{(m,m')}$, find all possible walks they can form, and assign each walk a relevance score. Note that a walk can have any length no larger than $L+1$. For a walk $[m_1,\cdots,m_a]$, 
we assign a relevance score in two way:
\begin{equation}
    \begin{aligned}
        \text{sum: } R^{[m_1,\cdots,m_a]} &= \sum_{i=1}^{a-1} R^{(m_i,m_{i+1})},\\
        \text{prod: } R^{[m_1,\cdots,m_a]} &= \prod_{i=1}^{a-1} R^{(m_i,m_{i+1})},
    \end{aligned}
    \notag
\end{equation}
which respectively correspond to Edge-IG sum and Edge-IG prod with $\widetilde{K}=25$ in the main text.
Note that these heuristics are compatible with all edge-level explanability methods, including GNNExplainer and PGExplainer, which however are  incomparably slow.

\section{More Samples for Qualitative Visualization}
\label{sec:AdditionalVisualization}
\subsection{Explaining model predictions on Mutagenicity}
\begin{figure}[H]
    \centering
    \includegraphics[width=\linewidth]{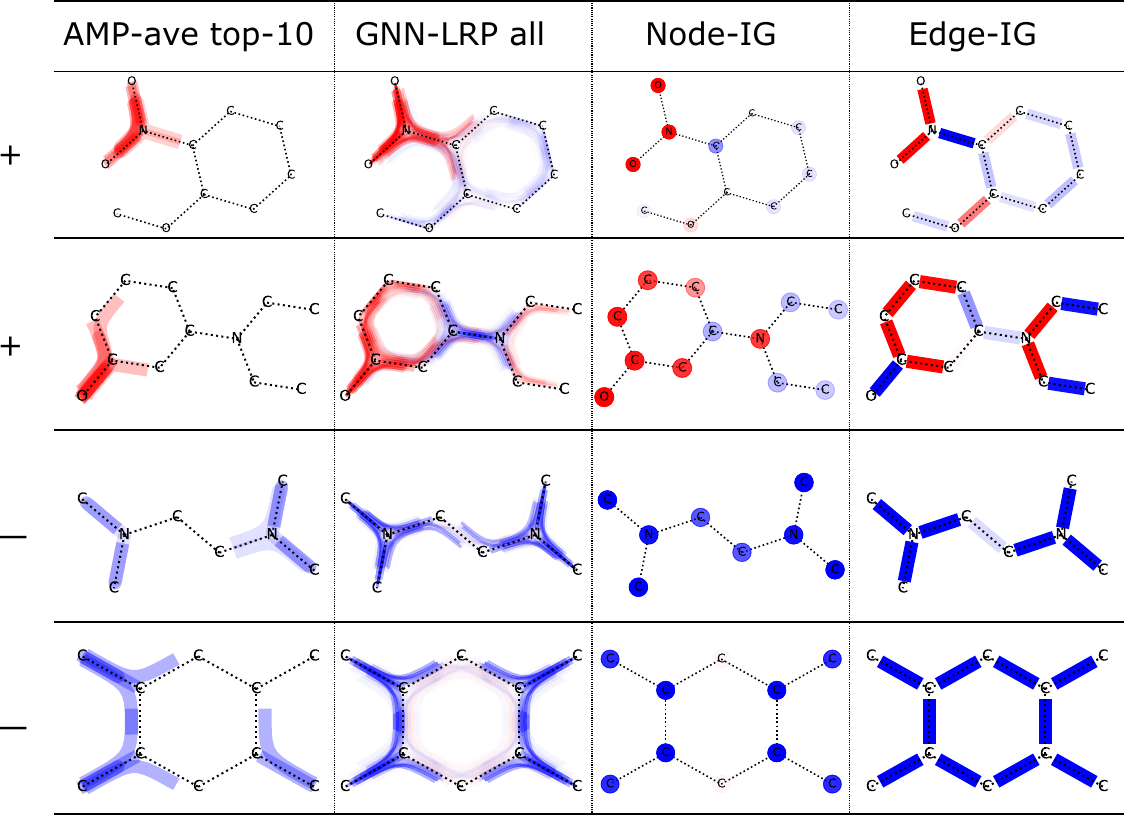}
    \caption{Heatmaps of different explanation methods of 4 molecules from the  Mutagenicity dataset (+: mutagenic, -: non-mutagenic).}
    \label{fig:vis-more-mol}
\end{figure}

In \cref{fig:vis-more-mol}, we give 4 further examples of molecule explanations, with two of them mutagenic and two non-mutagenic. 

In the first mutagenic molecule, our method successfully finds the mutagenic $\text{NO}_2$ group within the top-10 walks, while Node-IG and Edge-IG only highlight partially the $\text{NO}_2$ and wrongly give the carbon or C-N bond a negative relevance. GNN-LRP with all walks gives information of some non-mutagenic structures, but they are not as important as the mutagenic group that our method finds much faster.

In the second mutagenic molecule, our method identifies the OH substituent of the aromatic ring, while Edge-IG wrongly identifies the C-O bond as non-mutagenic. Node-IG is successful in identifying the real mutagenic structure, but wrongly identifies the nitrogen atom as mutagenic.

In the first non-mutagenic molecule, our method finds the C-N bonds, which are chemically stable and therefore indicate non-mutagenicity. Node-IG highlights all atoms, which is not wrong, but gives no specific information.

In the second non-mutagenic molecule, our method identifies the C subsituents of the aromatic rings, which lead to a chemically stable structure. However, Edge-IG highlights all edges and it's not clear if aromatic bond or the aromatic C structure is the evidence of non-mutagenicity.

\subsection{Explaining model predictions on Graph-SST2}
\begin{figure}[H]
    \centering
    \includegraphics[width=.8\linewidth]{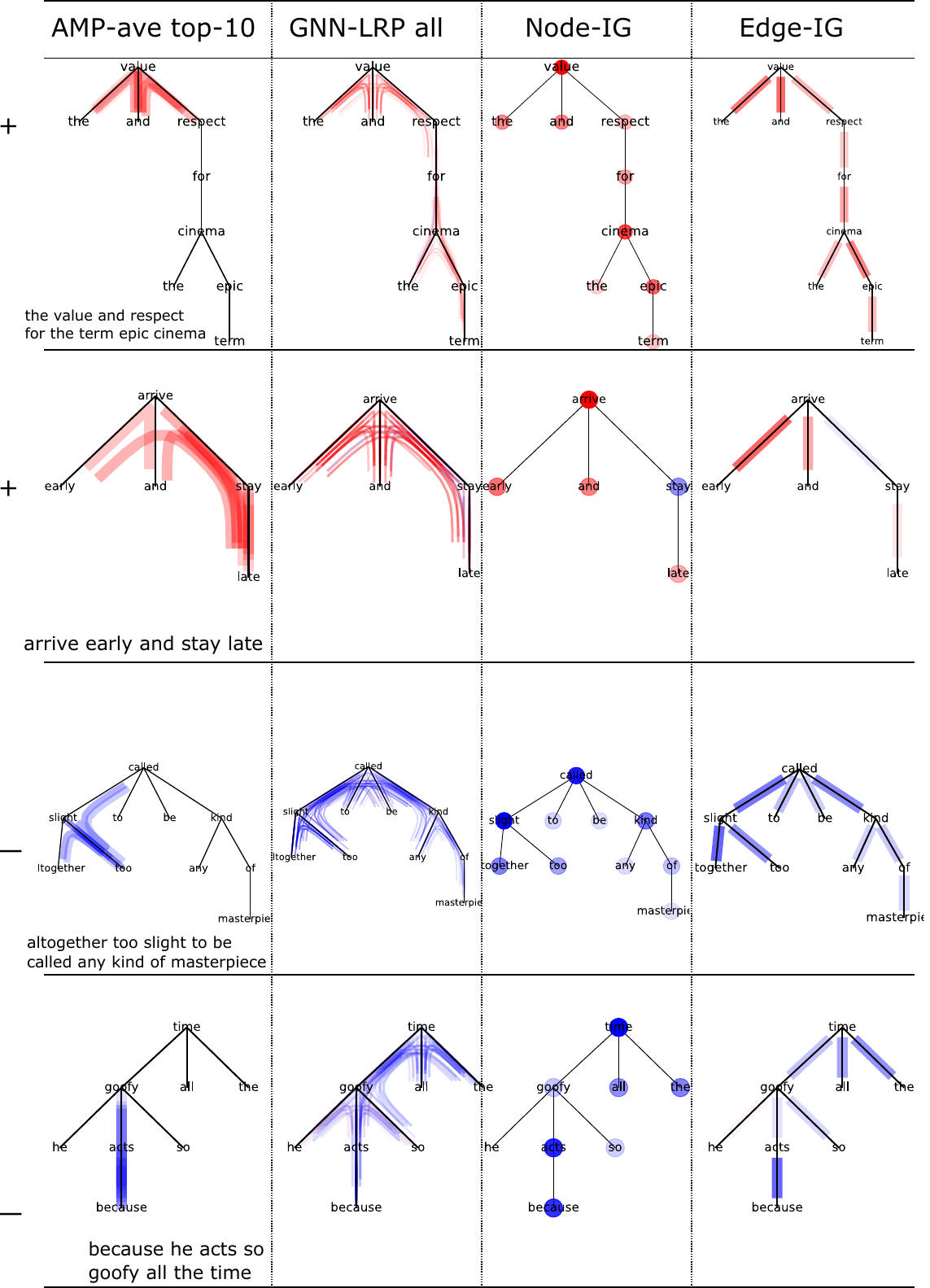}
    \caption{Explanations of parse trees from Graph-SST2 (+: positive sentiment, -: negative sentiment).}
    \label{fig:vis-more-sst}
\end{figure}

In \cref{fig:vis-more-sst}, we provide four further examples for parse tree explainations. Two of them with positive and two with negative sentiment.

In the first positive sentiment sample, our method finds the most relevant part ``the value and respect'', which is in this sentence most evident for the positive sentiment of the movie-review. When reading the sentence it is noticable that the last words ``epic cinema'' also give rise to a positive prediction, yet overall  less than ``the value and respect``. In the heatmap we see that our method decides for the most relevant subsentence and, with only 10 most relevant walks, leaves the less relevant parts out of scope.  Yet, we want to highlight that this is not a limitation of our method. Since, when we increase the number of top-$K$ walks, we would obtain other less relevant features in the parse-trees as well and finally converge with $K  \to \infty$ to the heatmap of GNN-LRP. We see that the baseline methods also provide a clear idea of what is relevant in the input graph, yet they are not able to distinguish comparably clear what is most relevant. 

In the second positive sentiment sample, we can say by reading the sentence that all words except ``and'' are relevant. From the heatmaps we see that our method identifies all of them with focus on ``arrive'' and ``stay late''. Notably, one of the top-10 walks goes through the four relevant words, showing our higher-order method's superiority in capturing the node interactions. On the opposite side, Node-IG wrongly identifies ``stay'' and Edge-IG wrongly identifies ``arrive''-``stay'' as evidence for negative sentiment.

In the first negative sentiment sample, our method finds the most relevant part ``altogether too slight'', which is what we expect to be evident for a negative sentiment. Node-IG highlights ``slight`` and ``called`` to be most evident, which is partly useful, but does not give sufficient intuition for this sentence. Edge-IG also highlights the edges between the words "altogether", "slight", "too" and "called", which seems reasonable for the prediction task. Yet, the connection between "called" and "kind" also seems quite relevant, which is more confusing here than enlightening. For the  GNN-LRP result we can see  that visualization all words can make it harder to tell which part of the sentence is more or less relevant, by just looking at the heatmaps. 

In the second negative sentiment sample, our method identifies the words ``because'' and ``acts goofy'', which are strong evidence for negative sentiment. If we plot all walks, the most highlighted part becomes ``goofy'' and ``time'', which is less intuitive for human. Node-IG and Edge-IG focus on the wrong parts, leaving the word ``goofy'' with only small relevance scores.

\end{document}